\documentclass[10pt,twocolumn,letterpaper]{article}

\usepackage[table,xcdraw,dvipsnames]{xcolor}
\usepackage[cvprfinal]{cvpr} %

\usepackage[dvipsnames]{xcolor}

\usepackage{makecell}
\usepackage[algo2e]{algorithm2e}
\usepackage{algorithm}
\usepackage{listings}
\usepackage{multirow}
\usepackage{soul}
\usepackage{wrapfig}

\newcommand{\fullmname}{CLIP as RNN}%
\newcommand{\mname}{CaR}%

\definecolor{cvprblue}{rgb}{0.21,0.49,0.74}
\usepackage[pagebackref,breaklinks,colorlinks,citecolor=cvprblue]{hyperref}

\def\paperID{1874} %
\def\confName{CVPR}
\def\confYear{2024}

\title{\fullmname: Segment Countless Visual Concepts without Training Endeavor}  %

\author{%
  Shuyang Sun$^{1, 2}$\footnotemark[1]\:
  \quad
  Runjia Li$^1$\footnotemark[1]\:
  \quad
  Philip Torr$^1$\:
  \quad
  Xiuye Gu$^2$\footnotemark[2]\:
  \quad
  Siyang Li$^2$\footnotemark[2]\\
  $^{1}$University of Oxford
  \quad
  $^2$Google Research
  \\
  \small
  \texttt{\{kevinsun, runjia, phst\}@robots.ox.ac.uk \{siyang, xiuyegu\}google.com}
  \\
  \url{https://torrvision.com/clip_as_rnn/}
}

\newcommand{\gain}[1]{\textcolor{Green}{+{#1}}}%
\newcommand{\tablestyle}[2]{\setlength{\tabcolsep}{#1}\renewcommand{\arraystretch}{#2}\centering\small}

\hypersetup{
    colorlinks=true,
    citecolor=RoyalBlue}

\begin{document}

\twocolumn[{
    \renewcommand\twocolumn[1][]{#1}
    \maketitle
    \begin{center}
    \captionsetup{type=figure}
    \includegraphics[width=\linewidth]{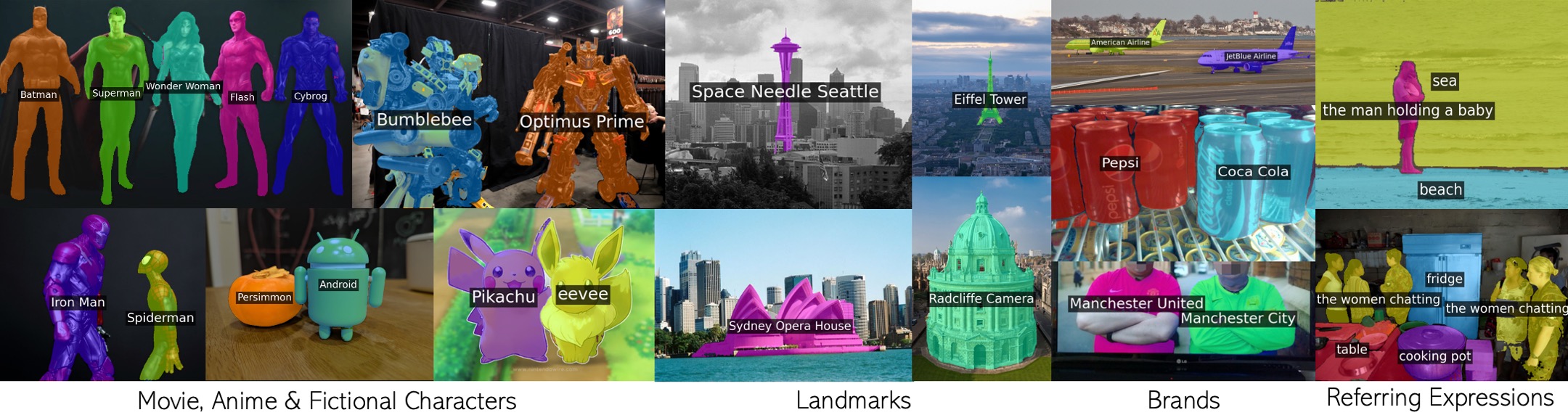}
    \caption{\textbf{We propose \textit{\fullmname} (\textit{\mname{}}) to segment concepts in a vast vocabulary,}
    including fictional characters, landmarks, brands, everyday objects, and referring expressions. This figure shows our qualitative results. More visualizations are included in the supplementary material. Best viewed in color and with zoom-in.
    }\label{fig:teaser}
    \end{center}
}]
\renewcommand{\thefootnote}{\fnsymbol{footnote}}
\footnotetext[1]{The first two authors contribute equally to this work. The majority of this work was done during Shuyang's internship at Google Research. }\footnotetext[2]{Equal advising.}
\begin{abstract}
\noindent Existing open-vocabulary image segmentation methods require a fine-tuning step on mask labels and/or image-text datasets. Mask labels are labor-intensive, which limits the number of categories in segmentation datasets. Consequently, the vocabulary capacity of pre-trained VLMs is severely reduced after fine-tuning. However, without fine-tuning, VLMs trained under weak image-text supervision tend to make suboptimal mask predictions.
To alleviate these issues, we introduce a novel recurrent framework that progressively filters out irrelevant texts and enhances mask quality without training efforts. The recurrent unit is a two-stage segmenter built upon a frozen VLM. Thus, our model retains the VLM's broad vocabulary space and equips it with segmentation ability.
Experiments show that our method outperforms not only the training-free counterparts, but also those fine-tuned with millions of data samples, and sets the new state-of-the-art records for both zero-shot semantic and referring segmentation. Concretely, we improve the current record by 28.8, 16.0, and 6.9 mIoU on Pascal VOC, COCO Object, and Pascal Context.
\end{abstract}

\section{Introduction}
Natural language serves as a bridge to connect visual elements with human-communicable ideas by transforming colors, shapes, and objects \etc into descriptive language. On the other hand, human can use natural language to easily instruct computers and robotics to perform their desired tasks.
Built upon the revolutionary vision-language model trained on Internet-scale image-text pairs, \eg, CLIP~\cite{clip},
a variaty of studies~\cite{liu2022open,xu2022groupvit,tcl,zhou2022extract,shin2022reco,segclip,clippy,ovseg,fcclip} have explored using pre-trained VLMs for \textit{open-vocabulary image segmentation} --- to segment any concepts in the image described by arbitrary text queries.

\begin{figure}
    \centering
    \tablestyle{0pt}{0.1}
    \begin{tabular}{ccc}
         \makecell[c]{OVSeg~\cite{ovseg}} & \makecell[c]{Grounded SAM~\cite{liu2023grounding}} &  \makecell[c]{\mname{} (Ours)}  \\
         \includegraphics[width=0.33\linewidth]{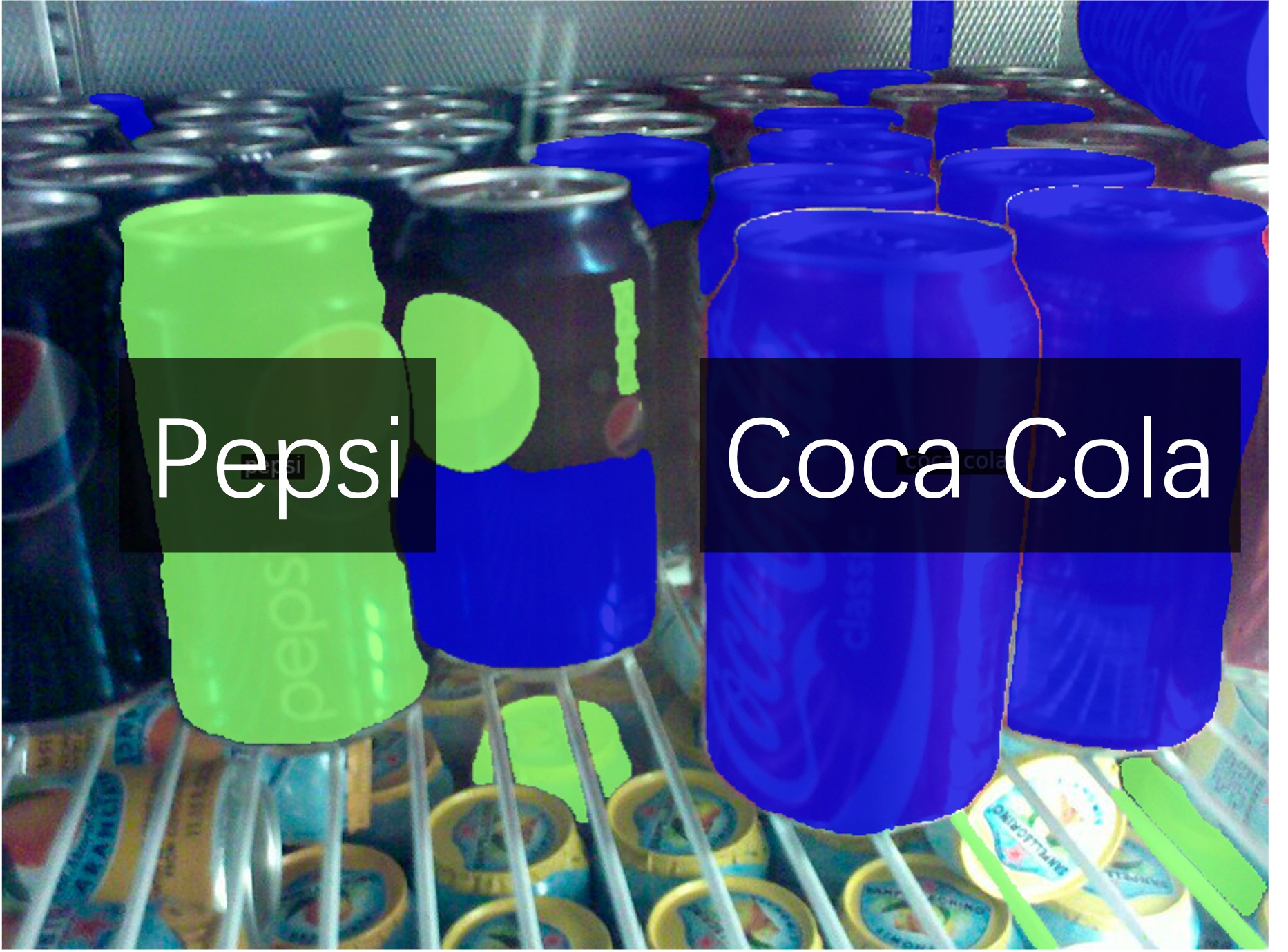} & \includegraphics[width=0.33\linewidth]{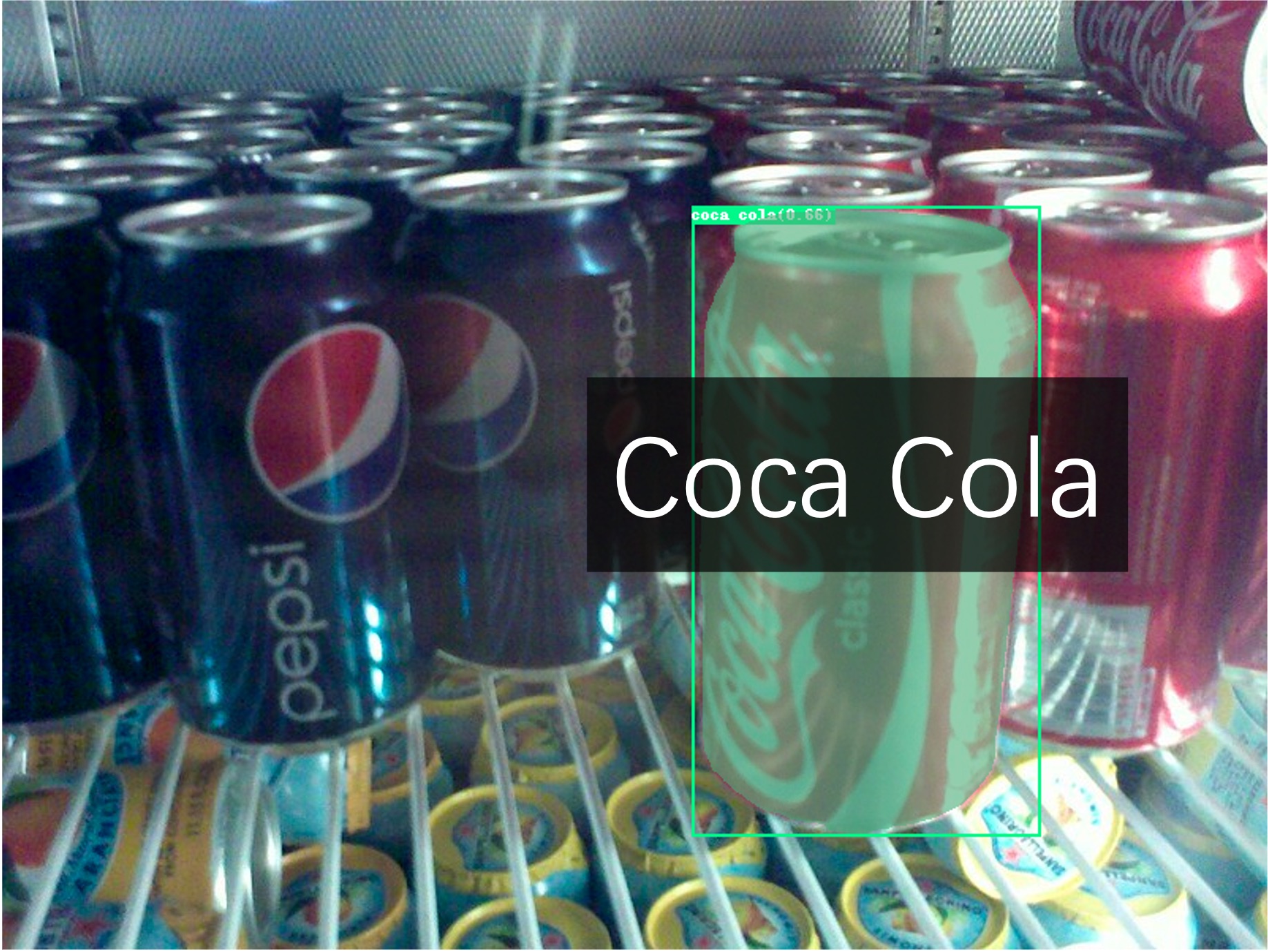} & \includegraphics[width=0.33\linewidth]{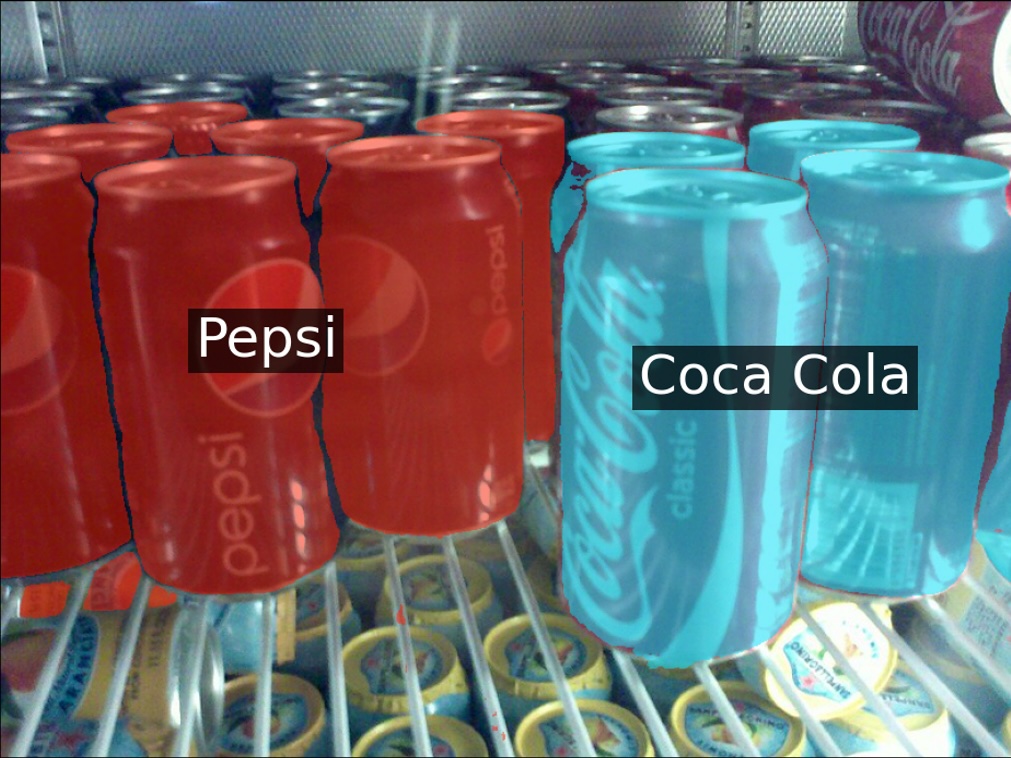}
    \end{tabular}
    \caption{\textbf{Our method \mname{} can fully inherit the vast vocabulary space of CLIP,} by directly using features from a pre-trained VLM, \ie, CLIP, without any fine-tuning. 
    Although the scene in the image is simple, state-of-the-art methods fine-tuned on segmentation datasets~\cite{liu2023grounding, ovseg} fail to segment and recognize \textit{Pepsi} and \textit{Coca Cola} correctly.}
    \label{fig:harry_potter}
\end{figure}

Among these advances, several works~\cite{ovseg, fcclip, liu2023grounding} have integrated pre-trained VLMs with segmenters fine-tuned on bounding boxes and masks. 
While these methods exhibit superior performances on segmentation benchmarks with common categories, their ability to handle a broader vocabulary is hampered by the small category lists in the segmentation datasets used for fine-tuning. %
As depicted in Figure~\ref{fig:harry_potter}, even though all three methods incorporate CLIP~\cite{clip}, those relying on fine-tuning with mask annotations~\cite{ovseg, liu2023grounding} fail to recognize the concepts like \textit{Pepsi} and \textit{Coca Cola}.

Since box and mask annotations are expensive, another line of works ~\cite{liu2022open,xu2022groupvit,tcl,ren2023viewco,segclip,clippy} seek to fine-tune the VLM and/or auxiliary segmentation modules with image-level annotations only, \eg, paired image-text data obtained from the Internet. This would lead to a complicated fine-tuning pipeline. Besides, these segmentation models often have suboptimal mask qualities, as image-level labels cannot directly supervise pixel grouping.

In this paper, we eliminate the fine-tuning on mask annotations or additional image-text pairs to fully preserve the extensive vocabulary space of the pre-trained VLM. However, the pre-training objectives of VLMs are not specifically designed for dense predictions. As a result, existing approaches~\cite{zhou2022extract,clipes,chen2022gscorecam} that do not fine-tune the VLMs struggle to generate accurate visual masks corresponding to the text queries, particularly when some of the text queries refer to non-existing objects in the image. To address this issue, we repeatedly assess the degree of alignment between mask proposals and text queries, and progressively remove text queries with low confidence. As the text queries become cleaner, better mask proposals are consequently obtained. To facilitate this iterative refinement, we propose a novel recurrent architecture with a two-stage segmenter as the recurrent unit, maintaining the same set of weights across all time steps. The two-stage segmenter consists of a mask proposal generator and a mask classifier to assess the mask proposals. Both are built upon a pre-trained CLIP model without modifications. Given an input image and multiple text queries, our model recurrently aligns the visual and textual spaces and generates refined masks as the final output, continuing until a stable state is achieved. Owing to its recurrent nature, we name our entire framework as \textit{\fullmname} (\textit{\mname}).

Experimental results demonstrate our approach is remarkably effective. In comparison with  methods that do not use additional training data,
\ie, zero-shot open-vocabulary semantic segmentation, our approach outperforms the prior art by {$28.8$}, {$16.0$}, and {$6.9$} mIoU on Pascal VOC~\cite{pascal_voc}, COCO Object~\cite{coco}, and Pascal Context~\cite{pascal_context}, respectively. Impressively, even when pitted against models fine-tuned on extensive additional data, our strategy surpasses the best record by {$12.6$}, {$4.6$}, and {$0.1$} on the three aforementioned datasets, respectively. To assess our model's capacity to handle more complex text queries, we evaluate on the referring image segmentation benchmarks, Ref-COCO, RefCOCO+ and RefCOCOg. \mname{} outperforms the zero-shot counterparts by a large margin.
Moreover, we extend our method to the video domain, and establish a zero-shot baseline for the video referring segmentation on Ref-DAVIS 2017~\cite{ref_davis}. 
As showcased in Figure~\ref{fig:teaser}, our proposed approach \mname{} exhibits remarkable success across a broad vocabulary spectrum, effectively processing diverse queries from celebrities and landmarks to referring expressions and general objects.

\noindent Our contributions can be summarized as follows:
 1. By constructing a recurrent architecture, our method \mname{} performs visual segmentation with arbitrary text queries in a vast vocabulary space without the need of fine-tuning. 
 2. When compared with previous methods on zero-shot open-vocabulary semantic segmentation and referring segmentation, our method \mname{} outperforms the prior state of the art by a large margin.

\begin{figure*}
    \centering
    \includegraphics[width=\linewidth]{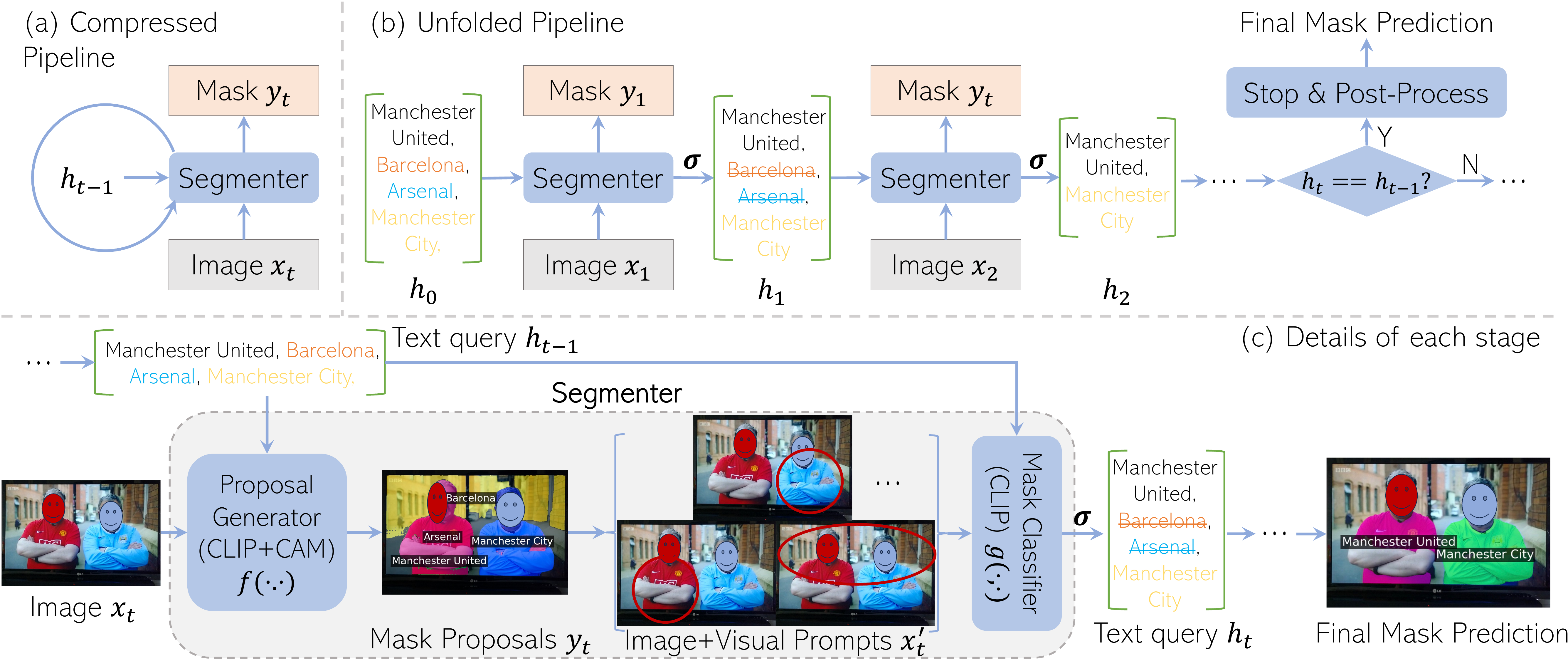}
    \caption{\textbf{The overall framework of our method \mname.}
    \textbf{(a)}, \textbf{(b)}: given an image, the user provides a set of text queries that they are interested to segment. This initial set, denoted by $h_0$, may refer to non-existing concepts in the image, \eg, \textcolor{Orange!80}{\texttt{Barcelona}} and \textcolor{Cyan}{\texttt{Arsenal}}. In the $t$-th time step, the frozen segmenter evaluates the degree of alignment between each mask and text query from the previous time step, $h_{t-1}$, and then low-confidence queries are eliminated by the function \( \sigma \).
    \textbf{(c)} depicts the detailed architecture of our two-stage segmenter. It consists a mask proposal generator \( f(\cdot, \cdot)\), and a mask classifier \( g(\cdot, \cdot)\) that assesses the alignment of each mask-text pairs.
    }
    \label{fig:method}
\end{figure*}

\section{Related Work}

\noindent\textbf{Open-vocabulary segmentation with mask annotations.}
The success of VLMs~\cite{clip, align, yu2022coca, zhai2022lit, flip, eva_clip, zhai2023sigmoid} has motivated researchers to push the boundaries of traditional image segmentation tasks, moving them beyond fixed label sets and into an open vocabulary by fine-tuning or training VLMs on segmentation datasets~\cite{zhang2022glipv2, owl_vit, liu2023grounding, openseg, fcclip, ovseg, odise, lseg, vild, detic, kamath2021mdetr, openseed}. However, as collecting mask annotations for a vast range of fine-grained labels is prohibitively expensive, existing segmentation datasets, \eg~\cite{coco, caesar2018coco, pascal_voc, pascal_context, zhou2019semantic} have limited vocabularies. Methods fine-tuned on these mask annotations reduce the open-vocabulary capacity inherited from the pre-trained VLMs. In this work, we attempt to preserve the completeness of the vocabulary space in pre-trained VLMs.

\noindent\textbf{Open-vocabulary segmentation without mask supervision.}
Several works~\cite{xu2022groupvit, zhou2022extract, shin2022reco, ren2023viewco, tcl, segclip, zeroseg, cai2023mixreorg, clippy, pacl, clips4, ovsegmenter} avoid the aforementioned vocabulary reduction issue by not fine-tuning on any mask annotations. Instead, researchers allow semantic grouping to emerge automatically without any mask supervision.
GroupViT~\cite{xu2022groupvit} learns to progressively group semantic regions with weak supervision, using only image-text datasets.
Furthermore, it is possible to use a pre-trained VLM for open-vocabulary segmentation without any additional training~\cite{zhou2022extract, shin2022reco, ovdiff}. For example, MaskCLIP~\cite{zhou2022extract} enables CLIP to perform open-vocabulary segmentation by only modifying its image encoder. However, these methods often suffer from inferior segmentation performance due to the lack of mask supervision, and the modification of the pre-trained VLMs. \mname{} is closely related to these approaches, we are both in a zero-shot manner without training. \mname{} stands out by proposing a recurrent framework on a VLM with fixed weights and no alternation on its architecture.
Note that our zero-shot is different from the zero-shot semantic segmentation~\cite{zs3net, spnet, hu2020uncertainty, zegformer, zhou2022extract,li2020consistent,baek2021exploiting} that mirrors the seen/unseen class separation from zero-shot classification in earlier ages.

\noindent\textbf{Segmentation with VLM-generated pseudo-labels.} 
As an alternative direction, recent works have exploited pre-trained VLMs to generate pseudo-masks in a fixed label space, requiring only image-level labels or captions for training~\cite{pamr, clipes, xie2022clims, mctformer, afa, clipseg, shin2022reco, zhou2022extract}. Once pseudo mask labels are obtained, a segmenter with a fixed vocabulary (\eg, DeepLab~\cite{deeplabv3plus2018, deeplabv12015}) can be trained in a fully supervised manner. Among these, CLIP-ES~\cite{clipes} is particularly relevant as it directly uses CLIP for pseudo-mask generation given the class names in ground-truth. However, CLIP-ES~\cite{clipes} requires pseudo-label training while we do not.

\noindent\textbf{Progressive refinement for image segmentation.}
Progressive refinement in image segmentation has seen significant advancements through various approaches. Recent works~\cite{detr, maxdeeplab, cheng2021per, cheng2021masked, yu2022kmax, sun2023remax} such as Cascade R-CNN~\cite{cai2018cascade}, DETR~\cite{detr} and CRF-RNN~\cite{crf_rnn} combine a detector (R-CNN~\cite{rcnn}), a transformer~\cite{transformer} or a segmenter (denseCRF~\cite{densecrf}) repeatedly for iterative refinement. We kindly note that all these works are designed for supervised image instance or semantic segmentation in a closed-set vocabulary. Our method does not require any training effort, yet our way of progressive refinement is fundamentally different from these methods.

\section{CLIP as Recurrent Neural Networks}

\subsection{A Recap on Recurrent Neural Networks}
We begin with a concise overview of recurrent neural networks (RNN). RNNs are specifically designed to process sequential data, such as text, speech, and time series. A basic RNN, commonly known as a vanilla RNN, uses the \textit{same} set of weights to process data at all time steps.
At each time step \( t \), the process can be expressed as follows:
\begin{align}
h_t & = \sigma(W_{hh} h_{t-1} + W_{xh} x_t + b_h), \\
y_t & = W_{hy} h_t + b_y.
\end{align}
\( x_t \) represents the input, and \( h_t \) represents the hidden state which serves as the ``memory" to store information from previous steps. \( y_t \) denotes the output. \( W_{hh} \), \( W_{xh} \), and \( W_{hy} \) are weight matrices, \( b_h \) and \( b_y \) refer to bias terms, and \( \sigma \) denotes a thresholding function to introduce non-linearity.

An RNN's core lies in its hidden state, \( h_t \), which captures information from past time steps. This empowers RNNs to exploit temporal dynamics within sequences. In our approach~\mname, we design a similar process - iteratively aligning the textual and visual domains by assessing the accuracy of each text query through a segmenter, {using the \textit{same} set of weights as well}. The text queries at each step act like the RNN's hidden state, representing the entities identified in the image at each specific time step.

\subsection{Overview}
\begin{algorithm}[t]
\caption{\small{Pseudo-code of CLIPasRNN in PyTorch style.}}
\definecolor{codeblue}{rgb}{0.25,0.5,0.5}
\definecolor{codegreen}{rgb}{0,0.6,0}
\definecolor{codegray}{rgb}{0.5,0.5,0.5}
\definecolor{codepurple}{rgb}{0.58,0,0.82}
\definecolor{backcolour}{rgb}{0.95,0.95,0.92}
\lstset{
  backgroundcolor=\color{white},
  basicstyle=\fontsize{7.2pt}{7.2pt}\ttfamily\selectfont,
  columns=fullflexible,
  breaklines=true,
  captionpos=b,
  commentstyle=\fontsize{7.2pt}{7.2pt}\color{codeblue},
  numberstyle=\tiny\color{codegray},
  stringstyle=\color{codepurple},
  keywordstyle=\fontsize{7.2pt}{7.2pt}\color{magenta},
}
\begin{lstlisting}[language=python]
# img: the input image with shape (3, H, W)
# h_0: a list of the initial N_0 text queries.
# clip: the CLIP model encoding the image and texts.
# cam: the gradient-based CAM model for mask proposal generation.
# eta: a threshold to binarize the masks for visual prompting.
# theta: a threshold defined in Eq. 6.

h_{t-1} = h_0
while len(h_{t-1}) > 0:
    # logits: [1, len(h_{t-1})]
    logits = clip(img, h_{t-1})
    scores = softmax(logits, dim=-1)
    # proposals: [len(h_{t-1}), H, W]
    proposals = cam(clip, img, scores)
    # prompted_img: [len(h_{t-1}), H, W]
    prompted_imgs = apply_visual_prompts(img, proposals, eta)
    # mask_logits: [len(h_{t-1}), len(h_{t-1})]
    mask_logits = clip(prompted_imgs, h_{t-1})
    mask_scores = softmax(mask_logits, dim=-1)
    # diag_scores: [len(h_{t-1})]
    diag_scores = diagonal(mask_scores)
    h_t = []
    for score, label in zip(diag_scores, h_{t-1}):
        if score > theta:
            h_t.append(label)
    if len(h_t) == len(h_{t-1}):
        break
    h_{t-1} = h_t
final_masks = post_process(proposals)

\end{lstlisting}
\label{algo}
\end{algorithm}

As depicted in Figure~\ref{fig:method}(a) and (b), our training-free framework operates in a recurrent manner, with a fixed-weight segmenter shared across all time steps.
In the $t$-th time step, the segmenter receives an image \( x_t \in \mathbb{R}^{3\times H\times W} \) and a set of text queries \( h_{t-1} \) from the preceding step as the input.
It then produces two outputs: a set of masks  \( y_t \in [0, 1]^{N_{t-1}\times H\times W} \) corresponding to $N_{t-1}$ input text queries, and the updated text queries $h_t$ for the subsequent step. For image segmentation, all different time steps share the same $x_t$.

To delve deeper into the design of our framework, we formulate its operations through Eq.~\eqref{eq:proposal} to Eq.~\eqref{eq:classifier}. %
\begin{align}\label{eq:proposal}
 y_t = f(x_t, h_{t-1}; W_f).
\end{align}
Here the function \(f(\cdot, \cdot)\) represents the mask proposal generator %
and $W_f$ denotes its pre-trained weights. The mask proposal generator processes the input image \(x_t\) and the text queries at previous step \( h_{t-1} \) to generate candidate mask proposals  \( y_t \).
Given the mask proposal generator is not pre-trained for dense prediction, the mask proposals \( y_t \) from \(f(\cdot, \cdot)\) are inaccurate. To assess these mask proposals, we draw visual prompts \eg, red circles or background blur, to the input $x_t$, based on mask proposals to highlight the masked area on the image. The visual prompting function $v(\cdot, \cdot)$ is defined as:
\begin{align}
    x^{\prime}_t = v(x_t, y_t).
    \label{eq:prompts}
\end{align}
Here \( x^{\prime}_t \) represent $N_{t-1}$ images with the visual prompts. The prompted images \( x^{\prime}_t \) are then passed to the mask classifier \( g(\cdot, \cdot) \) with the pre-trained weights $W_g$, along with the text queries \( h_{t-1} \), to compute a similarity matrix $P_t$. The entire process of the mask classifier can be defined as:
\begin{align}
    P_t = g(x^{\prime}_t, h_{t-1}; W_g).
    \label{eq:classifier}
\end{align}
Finally, after going through a thresholding function \( \sigma(\cdot) \), text queries with similarity scores lower than the threshold $\theta$ will be removed so that the text queries \( h_t = \sigma(P_t) \) for the next step $t$ are obtained. $h_{t}$ is a potentially reduced set of \( h_{t-1} \). Details of the thresholding function will be given in Section~\ref{sec:segmenter}.
This recurrent process continues until the text queries remain unchanged between consecutive steps, \ie, \(h_t == h_{t-1}\). We use $T$ to denote this terminal time step. Finally, we apply post-processing described in Section~\ref{sec:post_process} to the mask proposals $y_T$ generated in the final time step.

The pseudo-code in PyTorch-style is given in Algorithm~\ref{algo}. Note that users provide the initial text queries $h_0$, which are unrestricted and can include general object classes (``cat"), proper nouns (``Space Needle"), referring phrases (``the man in red jacket"), \etc.

\subsection{The Two-stage Segmenter}
\label{sec:segmenter}
In this section, we explain the two components of our segmenter, \ie a mask proposal generator and a mask classifier, which serve as the recurrent unit. As illustrated in Figure~\ref{fig:method}(c), the mask proposal generator first predicts a mask for each text query and then the mask classifier filters out irrelevant text queries based on the degree of alignment with their associated masks. 
We use the frozen pre-trained CLIP model weights for both the proposal generator and classifier to fully preserve the knowledge encapsulated in CLIP.

\begin{figure}
    \centering
    \tablestyle{0pt}{0.1}
    \begin{tabular}{ccccc}
    Red Circle & Red Contour & \makecell[c]{Background\\Blur} & \makecell[c]{Background\\Gray} & \makecell[c]{Background\\Mask} \\
      \includegraphics[width=0.2\linewidth]{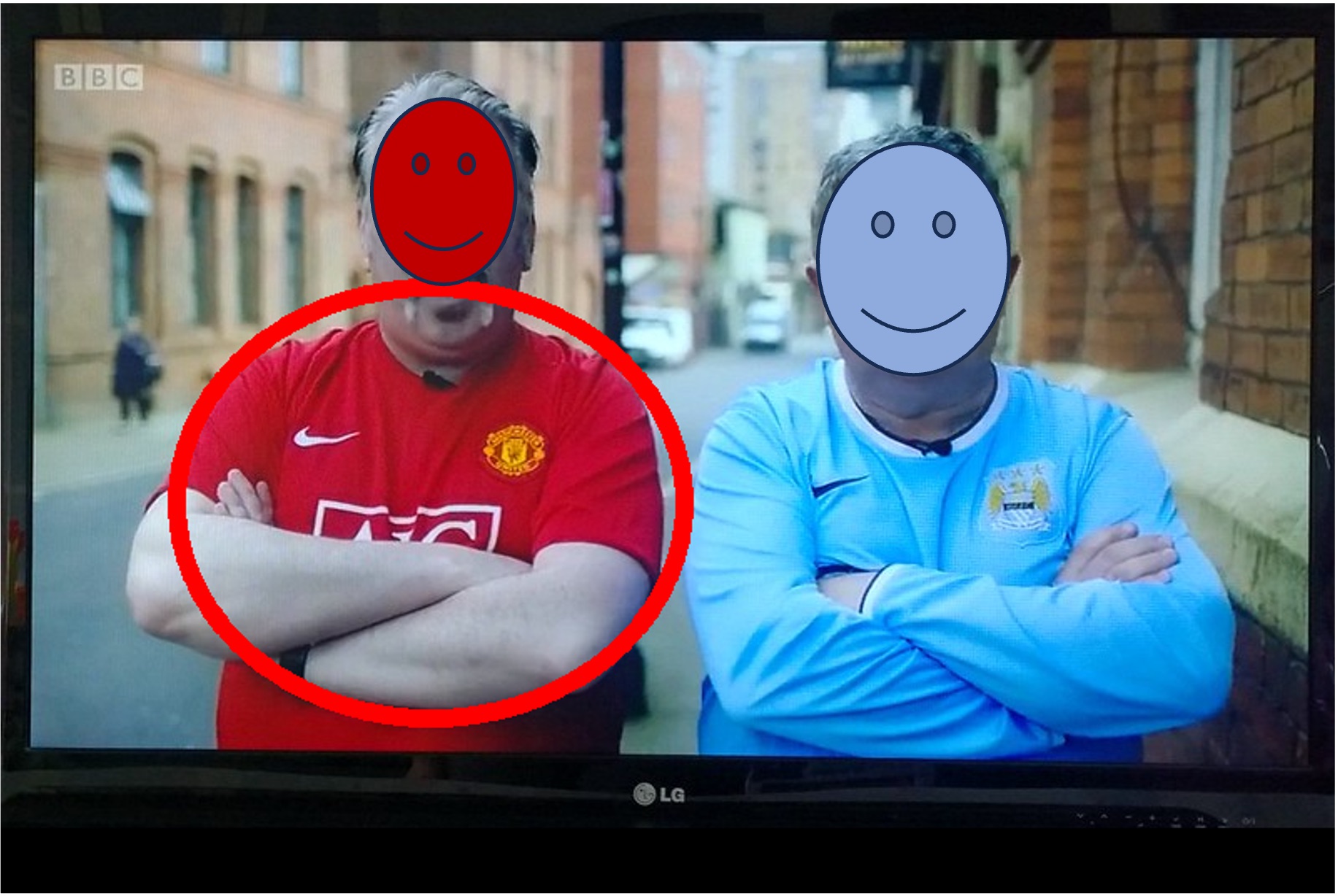} & \includegraphics[width=0.2\linewidth]{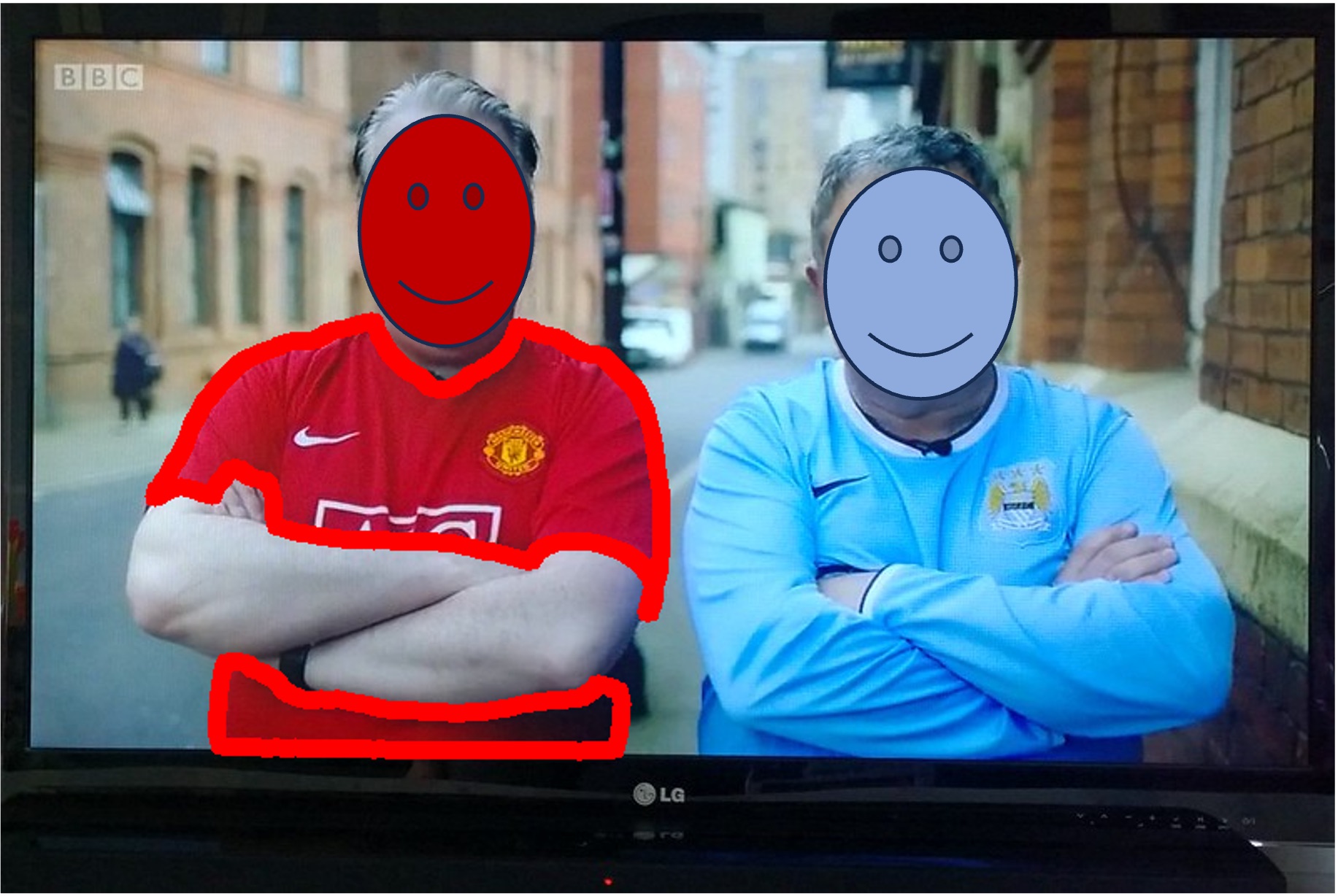} & \includegraphics[width=0.2\linewidth]{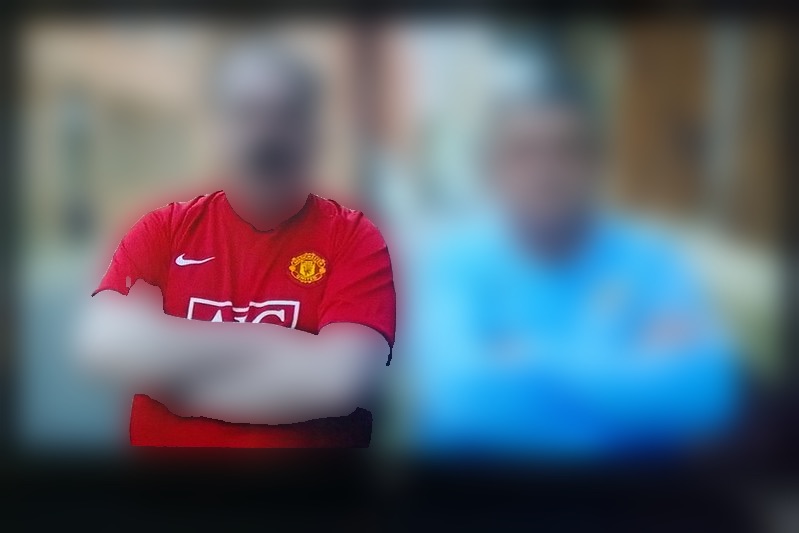} & \includegraphics[width=0.2\linewidth]{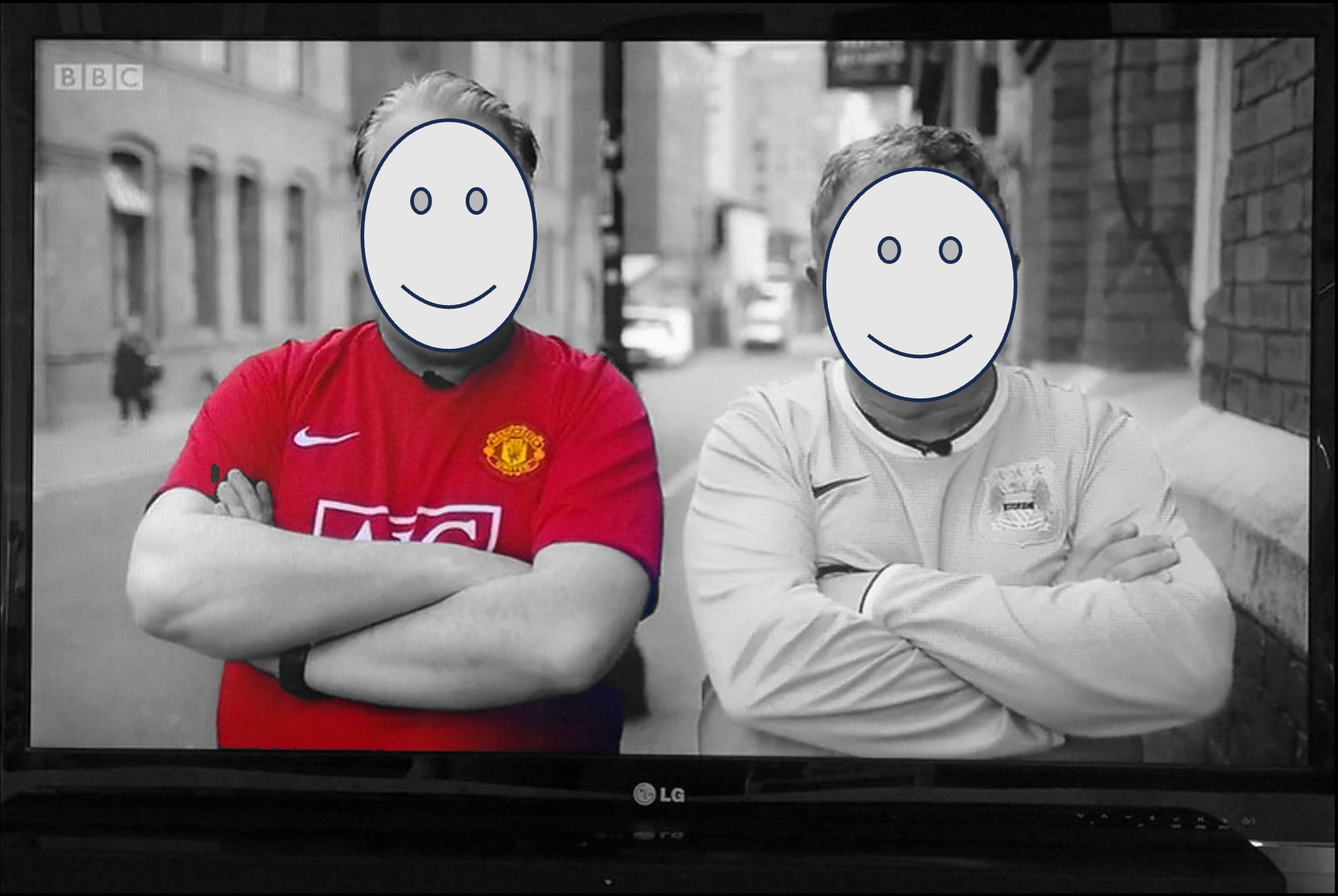} & \includegraphics[width=0.2\linewidth]{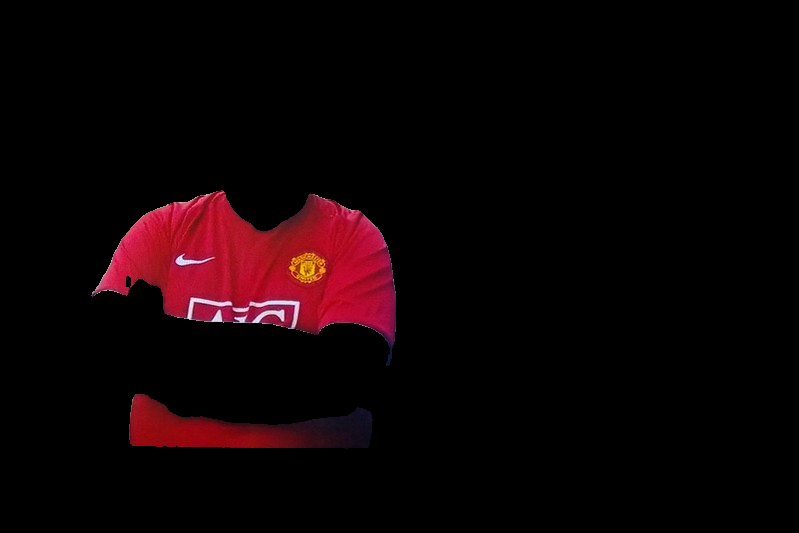}
    \end{tabular}
    \caption{Examples of visual prompts given a mask on the man wearing the jersey of Manchester United.}
    \label{fig:visual_prompts}
\end{figure}
\noindent\textbf{Mask proposal generator.} To predict the mask proposal $y_t$, a gradient-based Class-Activation Map (gradCAM)~\cite{gradcam, clipes} is applied to the pre-trained CLIP. More specifically, the image $x_t$ and text queries $h_{t-1}$ are first fed into CLIP to get a score between the image and each text. We then back-propagate the gradients of the score of each text query (\ie, class) from the feature maps of the CLIP image encoder to obtain a heatmap. Unless otherwise specified, we use the Class Activation Map (CAM) and class affinity (CAA) module from  CLIP-ES~\cite{clipes} as our mask proposal generator, with no further training of the CLIP model required. Apart from the text queries at the current step, we explicitly add a set of background queries describing categories that do not exist in the user text queries and calculate their gradients. This helps to suppress the activation from irrelevant texts (\eg, \textcolor{Orange!80}{\texttt{Barcelona}} and \textcolor{Cyan}{\texttt{Arsenal}} in Figure \ref{fig:method}) in the subsequent mask classification process.
More details of how CLIP works with gradCAM are provided in the appendix.

\noindent\textbf{Mask classifier.} The masks from the proposal generator may be noisy because the input texts are from an unrestricted vocabulary and may refer to non-existing objects in the input image.
To remove this type of proposals, we apply another CLIP model to compute a similarity score between each query and its associated mask proposal.
A straightforward approach is blacking out all pixels outside the mask region, as shown in the rightmost image in Figure~\ref{fig:visual_prompts}, and then computing the visual embedding for the foreground only. However, recent works~\cite{clipseg,red_circle_clip} have found several more effective \textit{visual prompts} which can highlight the foreground as well as preserve the context in the background. Inspired by this, we apply a variety of visual prompts, \eg, red circles, bounding boxes, background blur and gray background to guide the CLIP model to focus on the foreground region. A threshold $\eta$ is set to first binarize the mask proposals $y_t$ before applying these visual prompts to the images. Please refer to the supplementary material for more implementation details. After applying visual prompts, we obtain $N_{t-1}$ different prompted images, corresponding to $N_{t-1}$ text queries ($h_{t-1}$). We feed these images and text queries into the CLIP classifier $g(\cdot, \cdot)$ followed by a softmax operation along the text query dimension to get the similarity matrix $P_t \in \mathbb{R}^{N_{t-1} \times N_{t-1}}$ given the image and text embeddings.
We only keep the diagonal elements of $P_t$ as the matching score between the $i$-th mask and the $i$-th query. If the score is lower than a threshold $\theta$, the query and its mask are filtered out. Mathematically, the thresholding function $\sigma(\cdot)$ is defined as follows:
\begin{equation}
h_{t}^i = \sigma(P_t^{ii}) =
  \begin{cases}
    h_{t-1}^i, & \text{if } P_t^{ii} \geq \theta \\
    \texttt{NULL}, & \text{if } P_t^{ii} < \theta
  \end{cases}
  \label{eq:sigma}
\end{equation}
where $P_t^{ii}$ is the $i$-th element of the diagonal of the normalized similarity matrix, and $\theta$ is a manually set threshold. \texttt{NULL} represents that the $i$-th text query is filtered out and will not be input to next steps.

\begin{table*}[t!]
\centering 
\tablestyle{1.6pt}{1.0}
\scalebox{0.9}{
\begin{tabular}{lccc|cc|cccccccc}
\toprule
Models & \makecell[c]{Is \textit{VLM}\\ \textit{pre-trained?}} & \makecell[c]{\textit{w/ aux trainable}\\\textit{ module?}}  & \makecell[c]{Additional\\Training Data}  & \makecell[c]{\#Images}  & \makecell[c]{Additional\\Supervision}  &  \makecell[c]{VOC-\\20$^\ddagger$} & \makecell[c]{VOC-\\21} & \makecell[c]{COCO\\Object} & \makecell[c]{PC-\\59$^\ddagger$} & \makecell[c]{PC-\\60} & \makecell[c]{A-\\150} & \makecell[c]{A-\\847} & \makecell[c]{PC-\\459}  \\
\hline
\multicolumn{4}{l|}{\textit{zero-shot methods fine-tuned with additional data}} & & & & & & \\
  ViL-Seg \cite{liu2022open} & \checkmark & \checkmark  & CC12M & 12M & text+self & - & 34.4  & 16.4 & - & 16.3 & - & - & -\\
  GroupViT \cite{xu2022groupvit} & $\times$ & \checkmark  &  CC12M+YFCC & 26M & text & 74.1 & 52.3  &  24.3 & 23.4 & 22.4 & 10.6 & 4.3 & 4.9\\
  GroupViT \cite{xu2022groupvit}  & $\times$ & \checkmark  & CC12M+RedCaps & 24M & text  & 79.7 & 50.8  & 27.5 & - & 23.7 & - & - & - \\
 SegCLIP \cite{segclip} & $\times$ & \checkmark  & CC3M+COCO & 3.4M & text+self & - & 33.3  & 15.2 & - & 19.1 & - & - & -\\
 SegCLIP \cite{segclip} &  \checkmark & \checkmark  & CC3M+COCO & 3.4M & text+self & - & {52.6}  & 26.5 & - & 24.7 & 8.7 & - & - \\
 ZeroSeg \cite{zeroseg} & \checkmark & \checkmark  & IN-1K & 1.3M & self  & - & {40.8}  & {20.2} & - & {20.4} & - & - & - \\
 ViewCo \cite{ren2023viewco} & \checkmark & \checkmark  & CC12M+YFCC & 26M & text+self & - & 52.4  & 23.5 & - & 23.0 & - & - & -\\
 MixReorg~\cite{cai2023mixreorg} & \checkmark & \checkmark  & CC12M & 12M & text & - & 47.9 & - & - &  23.9 & - & - & -\\
 CLIPpy \cite{clippy} & \checkmark & $\times$  & HQITP-134M & 134M & text+self & - & 52.2  & \underline{32.0} & - & -  & 13.5 & - & -\\
 OVSegmenter \cite{ovsegmenter} & \checkmark & \checkmark  & CC4M & 4M & text & - & 53.8  & 25.1 & -& 20.4 & - & - & -\\
 TCL \cite{tcl} & \checkmark & \checkmark &  CC15M & 15M & text+self & 77.5 & {51.2} & 30.4 & 24.3 & {30.3} & 14.9 & - & -\\
 TCL+PAMR \cite{tcl} & \checkmark & \checkmark &  CC15M & 15M & text+self & \underline{83.2} & \underline{55.0} & 31.6 & \underline{33.9} & \underline{30.4} & \underline{17.1} & - & -\\
 \hline
 \multicolumn{4}{l|}{\textit{zero-shot methods with SAM}} & & & & & & \\
 SAMCLIP \cite{samclip} (\textit{w/} \cite{sam}) & \checkmark & \checkmark  & \footnotesize CC15M+YFCC+IN21k & 41M & text+self & - & 60.6 & - & - & 29.2 & 17.1 & - & - \\
 \mname+SAM (Ours, \textit{w/} \cite{sam_hq}) & \checkmark & - &  - & - & - & - & \textbf{70.2} & \textbf{37.6} & - & \textbf{31.1} & - & - & - \\  %
 \hline
 \multicolumn{4}{l|}{\textit{zero-shot methods without fine-tuning on CLIP}}  & & & & & & \\
 ReCo$^\dagger$ \cite{shin2022reco} & \checkmark & $\times$ & - & - & - & 57.7 & 25.1 & 15.7 & 22.3 & 19.9 & \underline{11.2} & - & -\\
  MaskCLIP$^\dagger$ \cite{zhou2022extract} & \checkmark & $\times$ & -  & - & - & \underline{74.9} & \underline{38.8} & \underline{20.6} & \underline{26.4} & \underline{23.6} & 9.8 & - & -\\
 \rowcolor{Gray!16}
 \mname{} (Ours, \textbf{\textit{w/o} SAM}) & \checkmark & $\times$ & - & - & - & \textbf{91.4} & \textbf{67.6}  & \textbf{36.6} & \textbf{39.5} & \textbf{30.5} & \textbf{17.7} & \textbf{8.1} & \textbf{13.9} \\  %
 \rowcolor{Gray!16}
 \multicolumn{3}{l}{$\Delta$ \textit{w/ the state-of-the-art w/o additional data}} & & & & \gain{16.5} & \gain{28.8} & \gain{16.0} & \gain{13.1} & \gain{6.9} & \gain{6.5} & - & -\\
 \rowcolor{Gray!16}
 \multicolumn{3}{l}{$\Delta$ \textit{w/ the state-of-the-art w/ additional data}} & & & & \gain{8.2} & \gain{12.6} & \gain{4.6} & \gain{5.6} & \gain{0.1} & \gain{0.6} & \gain{3.8} & \gain{9.0}\\

\bottomrule
\end{tabular}
}
\caption{
\textbf{Comparison to state-of-the-art zero-shot semantic segmentation approaches.} Results annotated with a $^\dagger$ are reported by \citet{tcl}. 
A $\checkmark$ is placed if either the visual or text encoder of the VLM is pre-trained. 
The table shows that our method outperforms not only counterparts without fine-tuning by a large margin, but also those fine-tuned on millions of data samples. For fair comparison, we compare with methods using CLIP~\cite{clip} as the backbone. $^\ddagger$VOC-20 and PC-59 represent Pascal VOC and Pascal Context \textbf{without} background. VOC-21 and PC-60 represent Pascal VOC and Pascal Context with an additional background class. A-150, A-847 and PC-459 represent ADE-150, ADE-847, and Pascal Context 459 datasets, respectively.
} 
\label{tab:semseg}
\end{table*}

\subsection{Post-Processing}
\label{sec:post_process}
Once the recurrent process stops, we start to post-process $y_T$, the masks from the final step $T$. We employ dense conditional random field (CRF)~\cite{densecrf} to refine mask boundaries. When constructing the CRF, the unary potentials are calculated based on the mask proposals of the last step. All hyper-parameters are set to the defaults in~\cite{densecrf}. Finally, an argmax operation is applied to the mask output of denseCRF along the dimension of text queries.
Thus, for each spatial location of the mask we only keep the class (text query) with the highest response. 

Additionally, we propose to ensemble the CRF-refined masks with SAM~\cite{sam}, as an \textbf{optional} post-processing module. This begins with generating a set of mask proposals from SAM using the \texttt{automask} mode, without entering any prompts into SAM. 
To match these SAM proposals with the masks processed by denseCRF, 
we introduce a novel metric: the \textit{Intersection over the Minimum-mask (IoM)}.
If the IoM between a mask from SAM and a CRF-refined mask surpasses a threshold $\phi_{iom}$, we consider them matched. Then all SAM proposals matched to the same CRF-refined mask are combined into one single mask. Finally, we compute the IoU between the combined mask and the original CRF-refined mask. If the IoU is greater than a threshold $\phi_{iou}$, we adopt the combined mask to replace the original mask, otherwise, we keep the CRF-refined mask.
The detailed post-processing steps are explained in the supplementary material.

\section{Experiments}
\subsection{Zero-shot Semantic Segmentation}
\noindent\textbf{Datasets.}
 Since our method does not require training, below we introduce only the datasets utilized for evaluation purposes. We conduct assessments for semantic segmentation using the validation (\texttt{val}) splits of Pascal VOC, Pascal Context, and COCO Object.
 Specifically, \textbf{Pascal VOC}~\cite{everingham2010pascal} encompasses 21 categories: 20 object classes (VOC-20) alongside one background class (VOC-21). For \textbf{Pascal Context}~\cite{pascal_context}, our evaluation employs the prevalent version comprising 59 classes including both ``things" and ``stuff" categories (PC-59), and one background (``other'') class for the concepts outside of the 59 classes (PC-60).
 Following~\cite{xu2022groupvit}, we construct the \textbf{COCO Object} dataset as a derivative of COCO Stuff~\cite{coco_stuff}. We kindly emphasize that the COCO Object dataset is \textbf{not} COCO Stuff since it merges all ``stuff" classes into one background class and thus has 81 classes (80 ``things" + 1 background) in total. We use the standard mean Intersection-over-Union (mIoU) metric to evaluate our method's segmentation performance. Besides, we report the performance of \mname{} on ADE-150 (A-150), ADE-847 (A-847) and Pascal Context 459 (PC-459) that consist of 150, 847 and 459 classes respectively.
 
\noindent\textbf{Implementation details.}
Our proposed method \mname{} utilizes the foundational pre-trained CLIP models as the backbone. More precisely, we harness the CLIP model with ViT-B/16 to serve as the underlying framework for the mask proposal generator $f(\cdot, \cdot)$. Concurrently, for the mask classifier $g(\cdot, \cdot)$, we adopt a larger ViT-L/14 version for higher precision based on our ablation study. Unless otherwise specified, the reported quantitative results are post-processed solely with a denseCRF, with no SAM masks involved. %
When setting the threshold hyper-parameters, we assign \(\eta=0.4\), \(\theta=0.6\), and \(\lambda=0.4\) for Pascal VOC, and \(\eta=0.5\), \(\theta=0.3\), \(\lambda=0.5\) for COCO and \(\eta=0.6\), \(\theta=0.2\), \(\lambda=0.4\) for Pascal context. 
The specific background queries used for the mask generator $f(\cdot, \cdot)$ are ablated in Section~\ref{sec:ablation} and detailed in the supplementary material. For Pascal Context, we use separate groups of background queries for ``thing" and ``stuff". For ``thing" categories, all ``stuff" categories are added as background queries and vice versa for ``stuff" categories.
As an optional strategy, we utilize a matching algorithm and perform an ensemble with SAM masks. We set both thresholds, $\phi_{iom}$ and $\phi_{iou}$, to 0.7 for all three datasets.
We enable half-precision floating point for CLIP. Since \mname{} is just a framework designed for inference, all experiments in this paper are conducted on just \textbf{one} NVIDIA V100 GPU.

\noindent\textbf{Efficiency.}  \mname{} \textbf{without SAM} operates at 950 ms with CPU-based CRF (200ms) for $500\times500$ images. GPU CRF is $\sim5\times$ faster. \mname{} is memory-efficient, consuming only 3.6GB GPU memory on Pascal VOC.

\begin{table}[t]
    \centering
    \tablestyle{6.6pt}{1.0}
    \begin{tabular}{c|cc|c}
    \toprule
    Dataset & \textit{w/ recurrence?} & CAM & mIoU \\
    \hline
        \multirow{3}{*}{Pascal VOC}& & CLIP-ES~\cite{clipes} & 15.2 \\
        & \checkmark & CLIP-ES~\cite{clipes} & 67.6 \\
        & \checkmark & gradCAM~\cite{gradcam} & 41.1 \\
    \bottomrule
    \end{tabular}
    \caption{\textbf{Effect of applying our recurrent architecure and different CAM methods.} The recurrence plays a vital role in improving the performance.}
    \label{tab:recurrence}
\end{table}

\noindent\textbf{\mname{} significantly outperforms methods without additional training.} 
We also compare \mname{} with training-free methods like MaskCLIP~\cite{zhou2022extract} and ReCo~\cite{shin2022reco}. 
Across the benchmarks, our model consistently demonstrates an impressive performance uplift. Under a similar setting with no additional training data, \mname{} surpasses previous state-of-the-art method by $28.8$, $16.0$ and $6.9$ mIoU on Pascal VOC, COCO Object and Pascal Context, respectively.

\noindent\textbf{Training-free \mname{} even outperforms several methods with additional fine-tuning.} 
As shown in Table~\ref{tab:semseg}, we compare our method with previous state-of-the-art methods including ViL-Seg~\cite{liu2022open}, GroupViT~\cite{xu2022groupvit}, SegCLIP~\cite{segclip}, ZeroSeg~\cite{zeroseg}, ViewCo~\cite{ren2023viewco}, CLIPpy~\cite{clippy}, and TCL~\cite{tcl}, which are augmented with additional data. The prior best results of different datasets are achieved by different methods. Specifically, TCL~\cite{tcl}, employing a fully pre-trained CLIP model and fine-tuned on 15M additional data, achieves the highest mIoU ($55.0$ and $30.4$) on Pascal VOC and Pascal Context. CLIPpy~\cite{clippy} sets the previous highest record on COCO Object but also requires extensive data for fine-tuning. Concretely, it first utilizes a ViT-based image encoder pre-trained with DINO~\cite{caron2021emerging} and a pre-trained T5 text encoder~\cite{t5}, and then fine-tunes both encoders with 134M additional data. Our method, incurring no cost for fine-tuning, still outperforms these methods by $12.6$, $4.5$, and $0.1$ mIoU on the Pascal VOC, COCO Object, and Pascal Context datasets, respectively. Since ``stuff'' appears less frequently in the pre-training image-text data for CLIP, \mname{} also exhibits less sensitivity to ``stuff'' on Pascal Context.

\noindent\textbf{\mname{}+SAM further boosts the performance.}
When integrated with SAM~\cite{sam, sam_hq}, we compare \mname~with a concurrent method SAMCLIP~\cite{samclip} and outperform it by $9.6$ and $1.9$ on Pascal VOC and Pascal Context.
We use the recent variant HQ-SAM~\cite{sam_hq} with \textbf{no prompt given} (\texttt{automask} mode), and then match the generated masks with metrics designed in Section~\ref{sec:post_process}. In other words, SAM is only used as a post-processor to refine the boundary of results from \mname. 
By applying SAM into our framework, our results can be further boosted by $2.6$, $1.1$ and $0.6$ mIoU on Pascal VOC, COCO Object and Pascal Context, respectively.

\begin{table}[t]
    \centering
    \tablestyle{6.6pt}{1}
    \begin{tabular}{cc|cc}
    \toprule
        \makecell[c]{Mask Proposal\\Generator \(f(\cdot, \cdot) \)} & \makecell[c]{Mask Classifier\\\(g(\cdot, \cdot) \)}&  \makecell[c]{Pascal\\ VOC} &  \makecell[c]{COCO\\ Object}\\
        \hline
        \multirow{2}{*}{ViT-B/16} & ViT-B/16 &  54.1 & 15.9\\  %
         & ViT-L/14 &  \textbf{67.6} & \textbf{36.6}\\ 
        \cline{1-4}
         \multirow{2}{*}{ViT-L/14} & ViT-B/16 &  50.6 & 14.1\\ %
         & ViT-L/14 & 57.6 & 32.5\\ %

        \bottomrule
    \end{tabular}
    \caption{\textbf{Effect of CLIP backbones.} We compare various CLIP backbones on Pascal VOC and COCO Object. Results show that we can improve the performance by scaling up the mask classifier.}
    \label{tab:backbone}
\end{table}
\begin{table}[t]
    \centering
    \tablestyle{3.6pt}{1.0}
    \begin{tabular}{c|ccccc|c}
    \toprule
        \multirow{2}{*}{Dataset} & \multicolumn{5}{c|}{Visual Prompts} & \multirow{2}{*}{mIoU} \\
         & \texttt{circle} & \texttt{contour} & \texttt{\makecell[c]{blur}} & \texttt{\makecell[c]{gray}} & \texttt{\makecell[c]{mask}} & \\
        \hline
        \multirow{10}{*}{\makecell[c]{Pascal\\VOC}} & \checkmark & & & & & 66.9 \\  %
        & & \checkmark & & & & {66.0} \\ %
        & & & \checkmark & & & 66.4 \\ %
        & & & & \checkmark & & 66.1 \\ %
        & & & & & \checkmark & {61.8} \\ %
        & \cellcolor{Gray!16}\checkmark & \cellcolor{Gray!16} & \cellcolor{Gray!16}\checkmark &
        \cellcolor{Gray!16} & \cellcolor{Gray!16} & \cellcolor{Gray!16}\textbf{67.6} \\ %
        & \checkmark & & & \checkmark & & 67.1\\ %
        & & \checkmark & \checkmark & & & 66.5 \\ %
        & & & \checkmark & \checkmark & & 66.3 \\ %
        & \checkmark & & \checkmark & \checkmark & & 66.8 \\ %
        \bottomrule
        
    \end{tabular}
    \caption{\textbf{Effect of different visual prompts.} When multiple visual prompts are checked, we will apply all checked visual prompts simultaneously on one image. The experiments are conducted on Pascal VOC. Results for COCO and Pascal Context are shown in supplementary materials.}
    \label{tab:visual_prompts}
\end{table}
\subsection{Ablation Studies.}
\label{sec:ablation}
\noindent\textbf{Effect of Recurrence.}
As illustrated in Table~\ref{tab:recurrence}, the incorporation of the recurrent architecture is crucial to our method. Without recurrence (\textit{a.k.a} $T=1$), our method functions similarly to CLIP-ES~\cite{clipes} with an additional CLIP classifier, and achieves only $15.2\%$ in mIoU. The recurrent framework can lead to a $52.4\%$ improvement, reaching an mIoU of $67.6\%$. The significant improvement validates the effectiveness of the recurrent design of our framework. For VOC and COCO, most images require two steps, and a small portion of images goes beyond two.

\noindent\textbf{Effect of different CAM methods.} Table~\ref{tab:recurrence} exhibits that our framework is compatible with different CAM methods and could be potentially integrated with other CAM-related designs. 
When integrated with CLIP-ES~\cite{clipes}, our method is 26.5 mIoU higher than that with gradCAM~\cite{gradcam}. We kindly note that we do not carefully search the hyper-parameters on gradCAM so the performance could be further improved.

\noindent\textbf{Effect of different CLIP Backbones.}
We experiment with different settings of CLIP backbones used in the mask proposal generator $f$ and mask classifier $g$, on Pascal VOC and COCO Object datasets.
Results are displayed in Table~\ref{tab:backbone}.
As for the mask proposal generator, ViT-B/16 outperforms the ViT-L/14 by over 10 mIoU on both Pascal VOC and COCO Object.
There is significant mIoU gains when employing the larger ViT-L/14 for the mask classifier over ViT-B/16. Similar observations have been found by \citet{red_circle_clip} that a larger backbone can better understand the visual prompts, which indicates that the performance of our method can be potentially improved by employing large backbones for the mask classifier.

\noindent\textbf{Effect of different visual prompts.}
There are various forms of visual prompts, including circle, contour, background blur (\texttt{blur}), background gray (\texttt{gray}), and background mask (\texttt{mask}), \etc.
We study the effects of different visual prompts on the Pascal VOC dataset and
Table~\ref{tab:visual_prompts} summarizes the results when applying one or a combination of two of the aforementioned visual prompt types.  
The highest mIoU score is achieved with the combination of \texttt{circle} and \texttt{blur}, yielding a mIoU of 67.6. Notably, using \texttt{mask} alone results in the lowest mIoU of 61.8, which is a common practice for most previous open-vocabulary segmentation approaches, \eg, \cite{ovseg, fcclip}.
We also evaluate the effect of different visual prompts on COCO Object and Pascal Context, and show the results in the supplementary material.

\begin{table}[t]
    \centering
    \tablestyle{3.6pt}{1.0}
    \begin{tabular}{cccc|cccc}
    \toprule
    \multicolumn{4}{c|}{Pascal VOC} & \multicolumn{4}{c}{COCO Object} \\
        $\eta$ & $\theta$ & $\lambda$ & {mIoU} & $\eta$ & $\theta$ & $\lambda$ & {mIoU} \\
        \hline
        0.3 & 0.6 & 0.4 & 67.0           & 0.5 & 0.3 & 0.6 & 35.4 \\
        0.4 & 0.6 & 0.4 & \textbf{67.6}  & 0.5 & 0.3 & 0.4 & 36.1 \\
        0.5 & 0.6 & 0.4 & 67.0           & 0.4 & 0.3 & 0.5 & 35.8 \\
        0.4 & 0.5 & 0.4 & 67.4           & 0.5 & 0.3 & 0.5 & \textbf{36.6} \\
        0.4 & 0.7 & 0.4 & 67.5           & 0.6 & 0.3 & 0.5 & 35.9 \\
        0.4 & 0.6 & 0.3 & 67.3           & 0.5 & 0.4 & 0.5 & 36.3 \\
        0.4 & 0.6 & 0.5 & 67.0           & 0.5 & 0.5 & 0.5 & 36.0 \\
        \bottomrule
    \end{tabular}
    \caption{\textbf{Effect of different hyper-parameters}: the threshold to binarize mask proposals ($\eta$), the threshold to remove text queries ($\theta$), and parameter of CLIP-ES's\cite{clipes} ($\lambda$). Experiments are conducted on Pascal VOC and COCO Object. }
    \label{tab:hyper_params}
\end{table}
\noindent\textbf{Effect of hyper-parameters.} %
We perform an ablation study on the performance impact of various hyper-parameter configurations on Pascal VOC, and present the results in Table~\ref{tab:hyper_params}. Hyper-parameters include the mask binarization threshold, \(\eta\), defined in Section \ref{sec:segmenter}, the threshold \(\theta\) employed in the thresholding function defined in Eq.~\eqref{eq:sigma}, and the parameter \(\lambda\) defined in CLIP-ES~\cite{clipes}. 
The peak performance is recorded at an mIoU of 67.6 for \(\eta=0.4\), \(\theta=0.6\), and \(\lambda=0.4\) on Pascal VOC and 36.6 for  \(\eta=0.5\), \(\theta=0.3\), and \(\lambda=0.5\) on COCO Object. Different parameter combinations result in mIoU scores that range from 67.0 to 67.6 on Pascal VOC and from 35.4 to 36.6 on COCO Object.

\begin{table}[t]
    \centering
    \tablestyle{1.6pt}{1.0}
    \begin{tabular}{c|ccc|c}
    \toprule
        \multirow{3}{*}{Dataset} & \multicolumn{3}{c|}{Background queries} & \multirow{3}{*}{mIoU} \\
         & \texttt{Terrestrial} & \makecell[c]{\texttt{Aquatic} \\ \texttt{Atmospheric}} & \makecell[c]{\texttt{Man-Made}} & \\
        \hline
        \multirow{8}{*}{\makecell[c]{Pascal\\VOC}} & $\times$ & $\times$ & $\times$ &  64.3 \\  %
        & \checkmark & $\times$ & $\times$ & 65.6 \\ %
        & $\times$ & \checkmark& $\times$ & 64.9 \\ %
        & $\times$ & $\times$ & \checkmark & 66.4 \\ %
        & \checkmark &\checkmark & $\times$ & 65.8\\ %
        & $\times$ &\checkmark & \checkmark & 66.4 \\ %
        &\checkmark & $\times$ & \checkmark &65.8 \\ %
         &\cellcolor{Gray!16}\checkmark & \cellcolor{Gray!16}\checkmark & \cellcolor{Gray!16}\checkmark & \cellcolor{Gray!16}\textbf{67.6}  \\ %
        \bottomrule
    \end{tabular}
    \caption{\textbf{Effect of background queries on Pascal VOC.} We divide background queries into: \texttt{Terrestrial}, \texttt{Aquatic}, \texttt{Atmospheric}, and \texttt{Man-Made}.
    We use ``None'' as the background query for the result in the first row.
    Specific background queries of each category are shown in the supplementary material.
    }
    \label{tab:bg_token}
\end{table}
\noindent\textbf{Effect of background queries.}
In Table~\ref{tab:bg_token}, we explore how different background queries (classes not existing in the input queries) can affect \mname{}'s performance.
We find that the segmentation quality improves as we include more diverse background queries: The combination of all three types of background queries delivers the highest mIoU of 67.6. For more details about the background queries of each class, please refer to the supplementary material.

\vspace{1ex}
\subsection{Referring Segmentation}
We evaluate \mname{} on the referring segmentation task for both images and videos. 
Again, our method is an inference-only pipeline built upon pre-trained CLIP models, and does not need training/fine-tuning on any types of annotations.
For referring segmentation we only use denseCRF~\cite{densecrf} for post-processing, and SAM is \textbf{not} involved for all experiments in this section for fair comparison.
Please refer to the supplementary material for the implementation details.

\begin{table}[t!]
\small 
\tablestyle{1.3pt}{1.0}
\scalebox{0.8}{
\begin{tabular}{lccc|ccc|ccc|c}
\toprule
Models  & \multicolumn{3}{c|}{RefCOCO} & \multicolumn{3}{c|}{RefCOCO+} & \multicolumn{3}{c|}{RefCOCOg} & GRES\\
 & val & testA & testB & val & testA & testB & val & test(U) & val(G) & \\
\hline
\multicolumn{4}{l|}{\textit{weakly-supervised}} & & & & & & & \\
TSEG \cite{tseg} & 25.95 & - & - & 22.62 & - & - & 23.41 & - & - & -\\
\hline
\multicolumn{4}{l|}{\textit{zero-shot}} & & & & & & \\
  \makecell[l]{GL CLIP} \cite{yu2023zerorefer} & 26.20 & 24.94 & 26.56 & 27.80 & 25.64 & 27.84 & 33.52 & 33.67 & 33.61 & -\\
 \mname (Ours)  & 33.57 & {35.36} & {30.51} & 34.22 & 36.03 & 31.02 & 36.67 & 36.57 & 36.63 & 16.8\\
\bottomrule
\end{tabular}
}
\caption{
\textbf{Comparison to state-of-the-art methods on referring image segmentation in mIoU.} \mname{} is better than all comparison methods in all benchmarks.
} 
\label{tab:refer_seg}
\end{table}

\noindent\textbf{Datasets.} 
Following~\cite{yang2022lavt, yu2023zerorefer}, we evaluate on \textbf{RefCOCO}~\cite{refcoco}, \textbf{RefCOCO+}~\cite{refcoco}, \textbf{RefCOCOg}~\cite{refcocog, refcocog_umd} and \textbf{GRES}~\cite{GRES} for the referring image segmentation task. Images used in all three datasets are sourced from the MS COCO~\cite{coco} dataset and {the masks} are paired with descriptive language annotations.
In RefCOCO+, descriptions about location are prohibited, making the task more challenging. There are two separate splits of the RefCOCOg dataset, one by UMD (U)~\cite{refcocog_umd} and another by Google (G)~\cite{wu2016google}. 
Following previous work, we use the standard mIoU metric.
Apart from referring image segmentation, we also set up a new baseline for \textbf{zero-shot referring video segmentation} on \textbf{Ref-DAVIS 2017}~\cite{ref_davis}. Following~\cite{ref_davis}, we adopt region similarity $\mathcal{J}$, contour accuracy $\mathcal{F}$, and the averaged score $\mathcal{J} \& \mathcal{F}$ as the metrics for evaluation.

\begin{wraptable}{r}{0pt}
    \centering
    \tablestyle{3.6pt}{1.0}
    \begin{tabular}{ccc}
    \toprule
        $\mathcal{J} \& \mathcal{F}$ & $\mathcal{J}$ & $\mathcal{F}$ \\
        \hline
        30.34 & 28.15 & 32.53 \\
    \bottomrule
    \end{tabular}
    \caption{\textbf{Results on Ref-DAVIS 2017.}}
    \label{tab:refdavis}
\end{wraptable}

\noindent\textbf{Experimental results.}
Table~\ref{tab:refer_seg} compares the performance of \mname{} with other methods on the referring image segmentation tasks across RefCOCO, RefCOCO+, and RefCOCOg. 
Comparing with other zero-shot methods, our method \mname{} outperforms Global-Local CLIP (GL CLIP) on all splits of these benchmarks. The performance gap is most pronounced on RefCOCO's \texttt{testA} split, where \mname{} outperforms 10.42 mIoU, and similarly on RefCOCO+'s \texttt{testA} split, with a lead of 10.72 mIoU. 
We also note that GL CLIP~\cite{yu2023zerorefer} uses a pre-trained segmenter Free-SOLO~\cite{wang2022freesolo} for mask extraction, while \mname{} is built \textbf{without} any pre-trained segmenter. As the first zero-shot method on GRES~\cite{GRES}, \mname{} achieves 16.8 mIoU with no training effort incurred.

For referring video segmentation, we demonstrate in Table~\ref{tab:refdavis} that our method achieves 30.34, 28.15 and 32.53 for $\mathcal{J} \& \mathcal{F}$, $\mathcal{J}$ and $\mathcal{F}$ on Ref-DAVIS 2017~\cite{ref_davis}. Considering that our method \mname{} requires neither fine-tuning nor annotations and operates in a zero-shot manner, this performance establishes a strong baseline.

\section{Conclusion}
\vspace{-1ex}
We introduce \fullmname~(\mname), which preserves the entire large vocabulary space of pre-trained VLMs, by eliminating the fine-tuning process. By constructing a recurrent pipeline with a shared segmenter in the loop, \mname{} can perform zero-shot semantic and referring segmentation without any additional training efforts. Experiments show that \mname{} outperforms previous state-of-the-art counterparts by a large margin on eight different benchmarks, \ie Pascal VOC, COCO Object, Pascal Context, ADE-150, ADE-847 and Pascal Context 459 on zero-shot semantic segmentation. We also demonstrate that \mname{} can handle referring expressions and segment fine-grained concepts like anime characters and landmarks, and also achieves state-of-the-art performance on RefCOCO, RefCOCO+, RefCOCOg and GRES for zero-shot referring segmentation. We hope our work sheds light on future research in open-vocabulary segmentation aiming to further expand the vocabulary space.

\vspace{1ex}
\noindent\textbf{Acknowledgement.}
The majority of this work was done during Shuyang’s internship at Google Research. We would like to thank Jindong Gu and Jianhao Yuan at Oxford University, Anurag Arnab, Xingyi Zhou, Huizhong Chen and Neil Alldrin at Google Research for their insightful discussion, Zhongli Ding for image donation.
Shuyang Sun and Philip Torr are supported by UKRI grants: Turing AI Fellowship EP/W002981/1. We would also like to thank the Royal Academy of Engineering and FiveAI.

{
    \small
    \bibliographystyle{ieeenat_fullname}
    \bibliography{main}
}

\appendix

\newpage
\clearpage

\section*{Appendix}
\renewcommand{\thetable}{\Alph{table}}
\renewcommand{\thefigure}{\Alph{figure}}
\renewcommand{\thealgorithm}{\Alph{algorithm}}

\section{More Experimental Results}
\subsection{Quantitative Analysis on Vocabulary Space.}
We demonstrate that our method \mname{} has a larger vocabulary space compared to the methods fine-tuned with mask annotations. Here we compare our method with OVSeg~\cite{ovseg}, which is fine-tuned on ImageNet~\cite{russakovsky2015imagenet} and COCO~\cite{coco} with a pre-trained CLIP backbone for the task of referring image segmentation.
We believe that referring expressions (\eg, ``the person in the red shirt'' or ``the cat in the mirror'') refers to a specific segment using a broad vocabulary. We conduct a comparative analysis between a robust open-vocabulary segmentation benchmark, OVSeg~\cite{ovseg}, and \mname, utilizing standard referring image segmentation benchmarks~\cite{refcoco,refcocog_umd, refcocog}. We note that RefCOCO and COCO share the same set of images so OVSeg fine-tuning on COCO may not be counted as zero-shot on RefCOCO.
The results, as detailed in Table \ref{tab:refer_ovseg}, demonstrate that \mname{} significantly surpasses OVSeg in performance. This disparity in performance suggests that \mname{} encompasses a more expansive vocabulary space than OVSeg.

\subsection{Evaluation without Background}
Following \cite{clipes}, our methodology benefits from using background queries in CLIP~\cite{clip} classification to suppress false positives (predictions not belonging to the input text queries), enhancing segmentation results. Nevertheless, for more comprehensive comparison, we also assess our approach using an alternate evaluation setting, previously established, which omits the background class. Consequently, less emphasis is placed on object boundaries in this setting. We test our method on two datasets: Pascal VOC~\cite{pascal_voc} without background (termed \textit{VOC-20}) and Pascal Context~\cite{pascal_context} without background (termed \textit{Context-59}). This setting tests the ability of various methods to discriminate between different classes. Our method \mname{} significantly outperforms previous methods on VOC-20 and Context-59, where all methods use the same setting that ignores the background class. 
We reached out to the PACL authors to confirm that they did not evaluate background.

\section{Implementation Details of CAM}
In this paper, we integrate two kinds of gradient-based CAM, \ie, Grad-CAM~\cite{gradcam} and CLIP-ES~\cite{clipes}, respectively.

\begin{table}[t!]
\small 
\tablestyle{1.6pt}{1.0}
\scalebox{0.86}{
\begin{tabular}{l|ccc|ccc|ccc}
\toprule
Models  & \multicolumn{3}{c|}{RefCOCO} & \multicolumn{3}{c|}{RefCOCO+} & \multicolumn{3}{c}{RefCOCOg}\\
 & val & testA & testB & val & testA & testB & val & test(U) & val(G) \\
\hline
  \makecell[l]{OVSeg} \cite{ovseg} & 22.58 & 19.38 & 25.63 & 19.13 & 15.74 & 25.30 & 27.87 & 29.09 & 28.31 \\
 \mname (Ours)  & 33.57 & {35.36} & {30.51} & 34.22 & 36.03 & 31.02 & 36.67 & 36.57 & 36.63\\
\bottomrule
\end{tabular}
}
\caption{
\textbf{Comparison to mask-supervised open-vocabulary methods on referring image segmentation in mIoU.} \mname{} is better than the comparison method, OVSeg, in all splits of the three benchmarks. 
}
\label{tab:refer_ovseg}
\end{table}
\begin{table}[t]
\centering
\tablestyle{1.6pt}{1.0}
\begin{tabular}{l|cc|cc}
\toprule
\multirow{2}{*}{Model} & \makecell[c]{{Is VLM}\\{pre-trained?}} & \makecell[c]{\#Additional\\Images} & \multicolumn{2}{c}{\textit{w/o} Background} \\
 & & & VOC-20 & Context-59  \\
\hline
GroupViT$^\dagger$ \cite{xu2022groupvit} & $\times$& 26M & 79.7 & 23.4  \\
PACL \cite{pacl} & \checkmark & 40M & 72.3 & 50.1  \\
TCL \cite{tcl} & \checkmark & 15M & 77.5 & 30.3  \\
\hline
MaskCLIP$^\dagger$ \cite{zhou2022extract} & \checkmark & - & 74.9 & 26.4  \\
ReCo$^\dagger$ \cite{shin2022reco} & \checkmark & - & 57.5 & 22.3 \\
\mname~(Ours) & \checkmark & - & \cellcolor{Gray!16}\textbf{91.4} & \cellcolor{Gray!16}\textbf{39.5} \\
\bottomrule
\end{tabular}
\caption{\textbf{Comparison with methods under the setting where background is ignored.} We compare \mname{} with prior work on VOC-20, Context-59 in a setting that considers only the foreground pixels (decided by ground truth). Our method shows comparable performance to prior works despite only relying on pretrained feature extractors. $^\dagger$: numbers are from \cite{tcl}.}
\label{table:your_label}
\end{table}

\paragraph{Integration with Grad-CAM.}
When integrating Grad-CAM~\cite{gradcam} into our framework, we first extract the image and text feature vectors \( v_x = f_I(x), v_h = f_T(h) \) from the image and text encoder \( f_I(\cdot), f_T(\cdot) \) given an image $x$ and text queries $h$. We compute a similarity score between the image and text features using the dot product \( s = \text{softmax}(v_x \cdot v_h^\intercal) \), where softmax is applied along the dimension of $v_h$. This score \( s \) quantifies the alignment (\textit{a.k.a} similarity) between the image \( x \) and the text \( h \) as perceived by the CLIP model. 
Here $h$ contains multiple queries. 
To integrate Grad-CAM into our framework, we first compute the gradients of the similarity score with respect to the feature maps of the image encoder by:
$$ g = \frac{\partial s}{\partial A^k}, $$
where \( A^k \) represents the feature maps and \(g\) denotes the gradients.
Then we compute the neuron importance weights by average-pooling the gradients:
 \[ \alpha_k = \frac{1}{Z} \sum_i \sum_j g_{ij}^k. \]
Here, \( \alpha_k \) is the neuron importance weights and \( Z \) is the number of pixels in each feature map. We then calculate a weighted combination of the feature maps \( A^k \) and the neuron importance \( \alpha_k \):
     \[ L = \text{ReLU}\left( \sum_k \alpha_k A^k \right), \]
an activation function ReLU is applied to filter out all negative activations. Specifically, we use the feature map after the first normalization layer of the last residual block to compute the gradients for CAM.

\paragraph{Integration with CLIP-ES.}
In summary, the CLIP-ES~\cite{clipes} we adopted is composed of a Grad-CAM and a class-aware attention-based affinity (CAA) module. The CAA module is introduced to enhance the vanilla multi-head self-attention (MHSA) for the Vision Transformer in CLIP. Given an image, class-wise CAM maps $M_c \in R^{h \times w}$ for each target class $c$ and the attention weight $W^{attn} \in R^{hw \times hw}$ are obtained from MHSA. For the attention weight, which is made asymmetric due to the use of different projection layers by the query and key, Sinkhorn normalization~\cite{sinkhorn1964relationship} is applied (alternately applying row-normalization and column-normalization) to convert it into a doubly stochastic matrix $D$, and the symmetric affinity matrix $A$ can be derived as follows:
\begin{equation}
  A = \frac{D + D^T}{2},  \text{ where } D = \text{Sinkhorn}(W^{attn}).
  \label{caa}
\end{equation}
For the CAM map $M_c \in R^{h \times w}$, a mask map for each target class $c$ can be obtained by thresholding the CAM with $\lambda$. Then a set of bounding boxes can be generated based on the thresholded masks. These boxes are used to mask the affinity weight matrix $A$, and then each pixel can be refined based on the masked affinity weight and its semantically similar pixels.
This refinement process can be formalized as follows:
\begin{equation}
  M_c^{aff} = B_c \odot A^t \cdot \text{vec}(M_c), \label{eq7}
\end{equation}
where $B_c \in R^{1 \times hw}$ represents the box mask obtained from the CAM of class $c$, $\odot$ denotes the Hadamard product, $t$ indicates the number of refining iterations, and $\text{vec}(\cdot)$ denotes the vectorization of a matrix. It should be noted that the attention map and CAM are extracted in the same forward pass. Therefore, CAA refinement is performed in real time and does not need an additional stage. Our implementation uses the attention maps from the last 8 layers of Vision Transformer for CAA.

\section{Implementation Details of Visual Prompts}
The Python code of visual prompts is shown in Algorithm~\ref{prompt_algo}, which is at the end of the supplementary material.
\begin{algorithm}[t]
\caption{\small{Pseudo-code of \fullmname{} in PyTorch style.}}
\definecolor{codeblue}{rgb}{0.25,0.5,0.5}
\definecolor{codegreen}{rgb}{0,0.6,0}
\definecolor{codegray}{rgb}{0.5,0.5,0.5}
\definecolor{codepurple}{rgb}{0.58,0,0.82}
\definecolor{backcolour}{rgb}{0.95,0.95,0.92}
\lstset{
  backgroundcolor=\color{white},
  basicstyle=\fontsize{7.2pt}{7.2pt}\ttfamily\selectfont,
  columns=fullflexible,
  breaklines=true,
  captionpos=b,
  commentstyle=\fontsize{7.2pt}{7.2pt}\color{codeblue},
  numberstyle=\tiny\color{codegray},
  stringstyle=\color{codepurple},
  keywordstyle=\fontsize{7.2pt}{7.2pt}\color{magenta},
}
\begin{lstlisting}[language=python]
import cv2
import numpy as np
import torch
from scipy.ndimage import binary_fill_holes
def apply_visual_prompts(
        image_array,
        mask,
        visual_prompt_type=('circle'),
        visualize=False,
        color=(255, 0, 0),
        thickness=1,
        blur_strength=(15, 15)):
    prompted_image = image_array
    inv_mask = (1 - mask)[:, :, None]
    if 'blur' in visual_prompt_type:
        # blur the part out side the mask
        # Blur the entire image
        blurred = cv2.GaussianBlur(prompted_image, blur_strength, 0)
        # Get the sharp region using the mask
        sharp_region = cv2.bitwise_and(
            prompted_image,
            prompted_image,
            mask=np.clip(mask, 0, 255)))
        # Get the blurred region using the inverted mask
        
        blurred_region = (blurred * inv_mask)
        # Combine the sharp and blurred regions
        prompted_image = cv2.add(sharp_region, blurred_region)
    if 'gray' in visual_prompt_type:
        gray = cv2.cvtColor(prompted_image, cv2.COLOR_BGR2GRAY)
        # make gray part 3 channel
        gray = np.stack([gray, gray, gray], axis=-1)
        # Get the sharp region using the mask
        color_region = cv2.bitwise_and(
            prompted_image,
            prompted_image,
            mask=np.clip(mask, 0, 255))
        # Get the blurred region using the inverted mask
        inv_mask = 1 - mask
        gray_region = (gray * inv_mask)
        # Combine the sharp and blurred regions
        prompted_image = cv2.add(color_region, gray_region)
    if 'black' in visual_prompt_type:
        prompted_image = cv2.bitwise_and(
            prompted_image,
            prompted_image,
            mask=np.clip(mask, 0, 255))
    if 'circle' in visual_prompt_type:
        mask_center, mask_height, mask_width = mask2chw(mask)
        center_coordinates = (mask_center[1], mask_center[0])
        axes_length = (mask_width // 2, mask_height // 2)
        prompted_image = cv2.ellipse(prompted_image,
                                     center_coordinates,
                                     axes_length, 0, 0, 360, color, thickness)
    if 'rectangle' in visual_prompt_type:
        mask_center, mask_height, mask_width = mask2chw(mask)
        center_coordinates = (mask_center[1], mask_center[0])
        start_point = (mask_center[1] - mask_width //
                       2, mask_center[0] - mask_height // 2)
        end_point = (mask_center[1] + mask_width //
                     2, mask_center[0] + mask_height // 2)
        prompted_image = cv2.rectangle(prompted_image,
                                       start_point,
                                       end_point,
                                       color, thickness)
    if 'contour' in visual_prompt_type:
        # Find the contours of the mask
        # fill holes for the mask
        mask = binary_fill_holes(mask)
        contours, hierarchy = cv2.findContours(mask, cv2.RETR_TREE, cv2.CHAIN_APPROX_SIMPLE)
        # Draw the contours on the image
        prompted_image = cv2.drawContours(
            prompted_image, contours, -1, color, thickness)
    return prompted_image

\end{lstlisting}
\label{prompt_algo}
\end{algorithm}

\section{Breakdown of Background Tokens}
We break down the background tokens into 3 sub-categories for ablation study (experiment results are shown in the main manuscript in Table 6):
\begin{itemize}
    \item \texttt{Terrestrial}: [`ground', `land', `grass', `tree',  `mountain', `rock', `valley', `earth',`terrain', `forest', `bush', `hill', `field', `pasture', `meadow', `plateau', `cliff', `canyon', `ridge', `peak', `plain', `prairie', `tundra', `savanna', `steppe', `crag', `knoll', `dune', `glen', `dale', `copse', `thicket']
    \item \texttt{Aquatic-Atmospheric}: [`sea', `ocean', `lake',  `water', `river', `sky', `cloud', `pond', `stream', `lagoon', `bay', `gulf', `fjord', `estuary', `creek', `brook', `reservoir', `pool', `spring', `marsh', `swamp', `wetland', `glacier', `iceberg', `atmosphere', `stratosphere', `mist', `fog', `rain', `drizzle', `hail', `sleet', `snow', `thunderstorm', `breeze', `wind', `gust', `hurricane', `tornado', `monsoon',  `cumulus', `cirrus', `stratus', `nimbus']  
    \item \texttt{Man-Made}: [
      `building', `house', `wall', `road', 
      `street', `railway', `railroad', `bridge',
      `edifice', `structure', `apartment', `condominium', `skyscraper', `highway', `boulevard', `lane', `alley', `byway', `avenue', `expressway', `freeway', `path', `overpass', `underpass', `viaduct', `tunnel', `footbridge', `crosswalk', `culvert', `dam', `archway', `causeway', `plaza', `square', `station', `terminal'
  ]
\end{itemize}

\section{Implementation Details of Mutual Background for Pascal Context}
Our approach involves creating a list of background queries to minimize false positive predictions in mask proposals. However, in the Pascal Context dataset \cite{pascal_context}, many ``stuff'' categories (\eg sky, ground, sea) serve as background queries for ``object'' categories (\eg bird, car, boat). Directly removing these 'stuff' categories from the background query list and generating object and stuff masks using CAM leads to noisy results due to the lack of false positive background suppression. To address this issue, we adopt a mutual background strategy. In this method, object and stuff masks are produced separately, using object categories as the background queries for stuff masks and vice versa. This technique not only maintains the benefit of reducing false positives but also significantly enhances performance in the Pascal Context dataset.

\section{Implementation Details of Referring Image Segmentation.}
We use ViT-B/16 as the backbone of the visual encoder for both the mask proposal generator and mask classifier, and use \texttt{circle} and \texttt{background blur} as the visual prompts for the inputs of mask classifier. The $\eta$, $\theta$, $\lambda$ were set to (0.5, 0.3, 0.5), (0.2, 0.1, 0.5), (0.5, 0.1, 0.6) for refCOCO, refCOCO+ and refCOCOg, respectively. All splits of these three datasets share the same set of hyper-parameters. We note that we \textbf{do not} apply SAM for referring image segmentation.

\section{More Visualization Results}

\subsection{Visualization results on different post-processors}
\begin{figure*}
    \centering
    \tablestyle{0pt}{0.1}
    \begin{tabular}{cccc}
      Image & \mname & \mname+SAM & GT \\
      \includegraphics[width=0.25\linewidth]{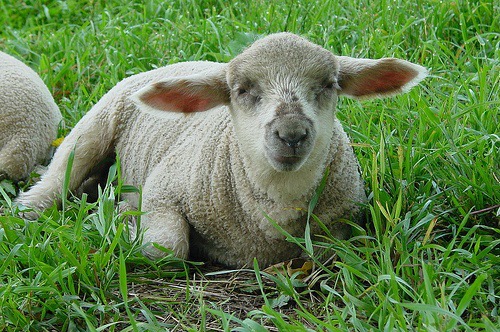} & \includegraphics[width=0.25\linewidth]{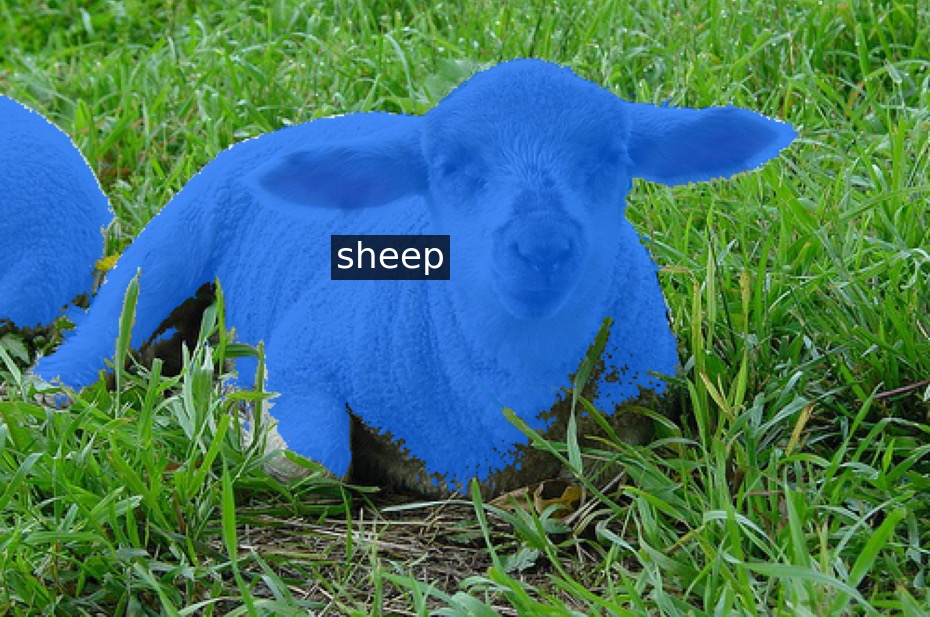} & \includegraphics[width=0.25\linewidth]{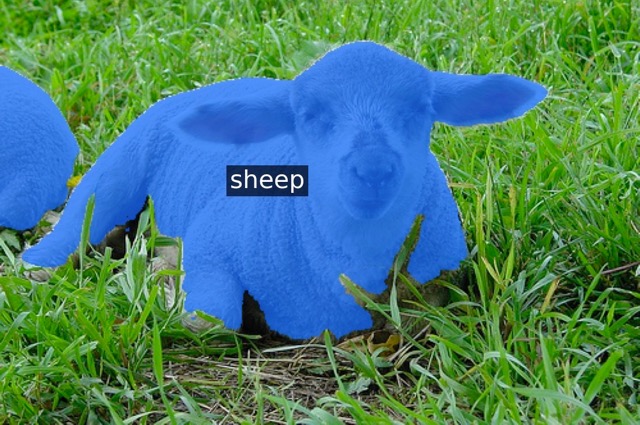} & \includegraphics[width=0.25\linewidth]{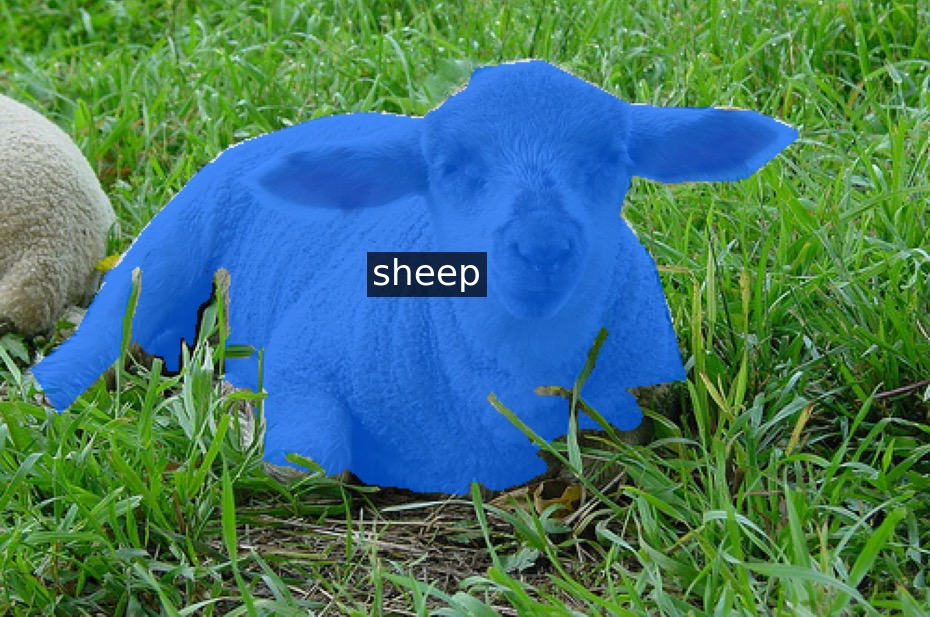}  \\
        \includegraphics[width=0.25\linewidth]{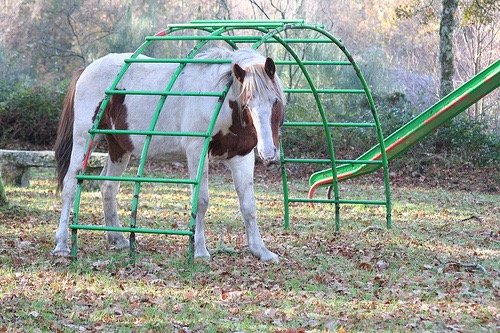} & \includegraphics[width=0.25\linewidth]{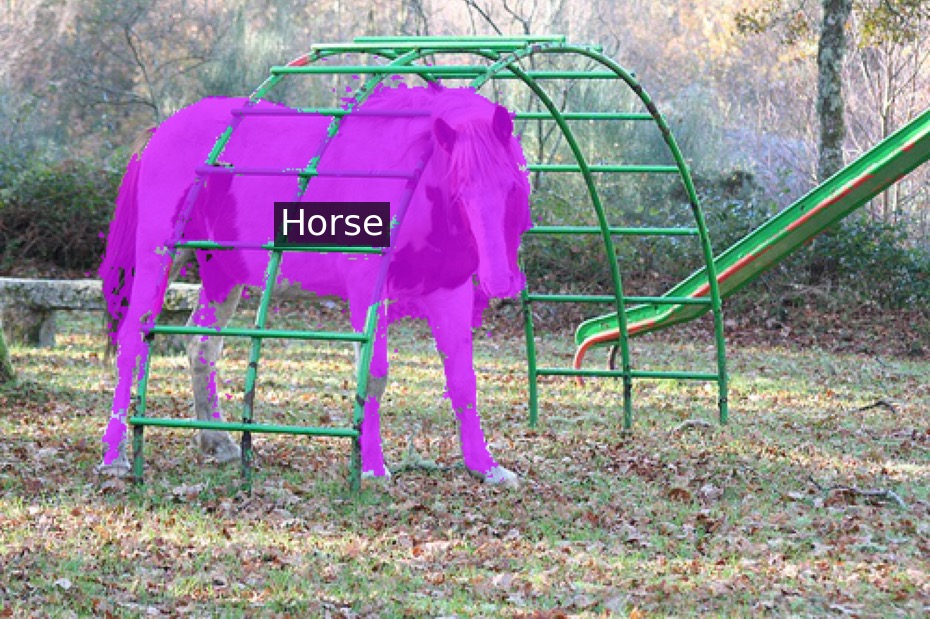} & \includegraphics[width=0.25\linewidth]{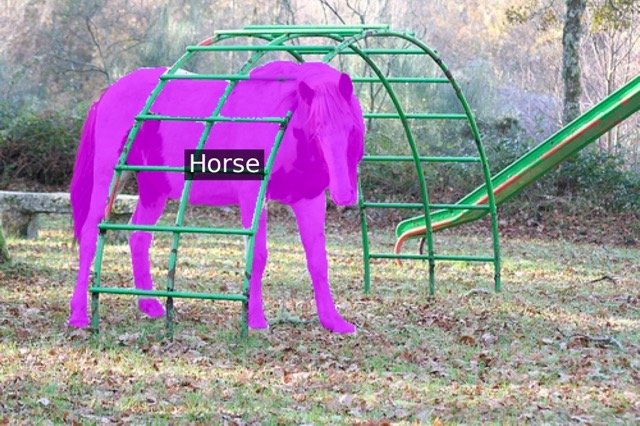}& \includegraphics[width=0.25\linewidth]{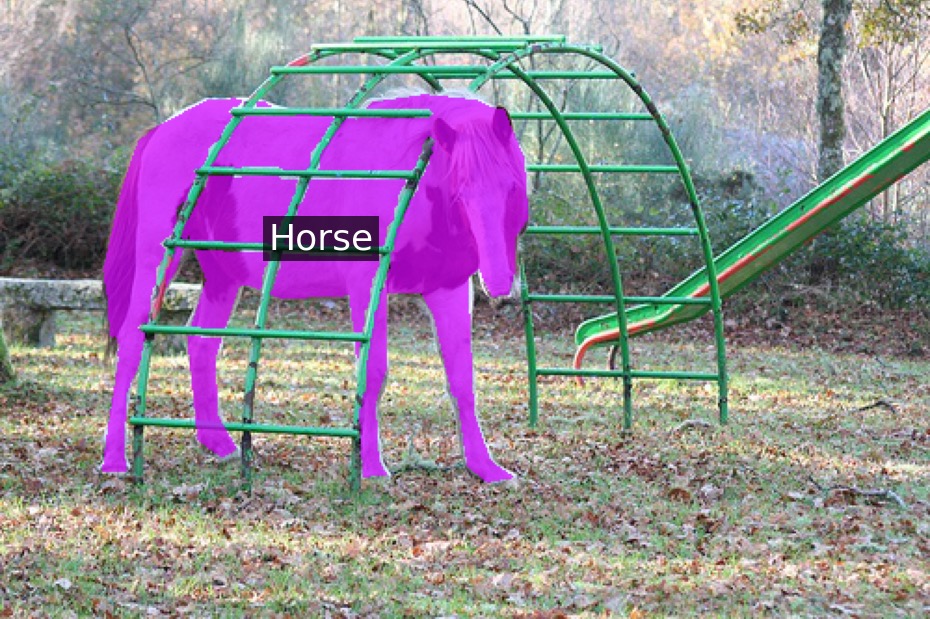} \\
      \includegraphics[width=0.25\linewidth]{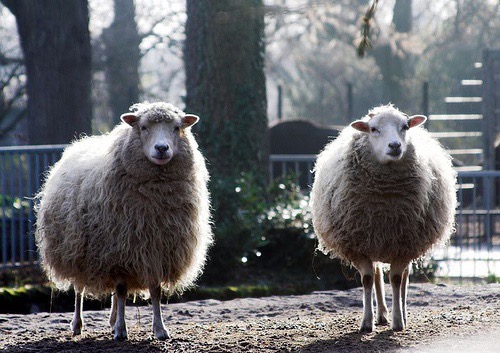} & \includegraphics[width=0.25\linewidth]{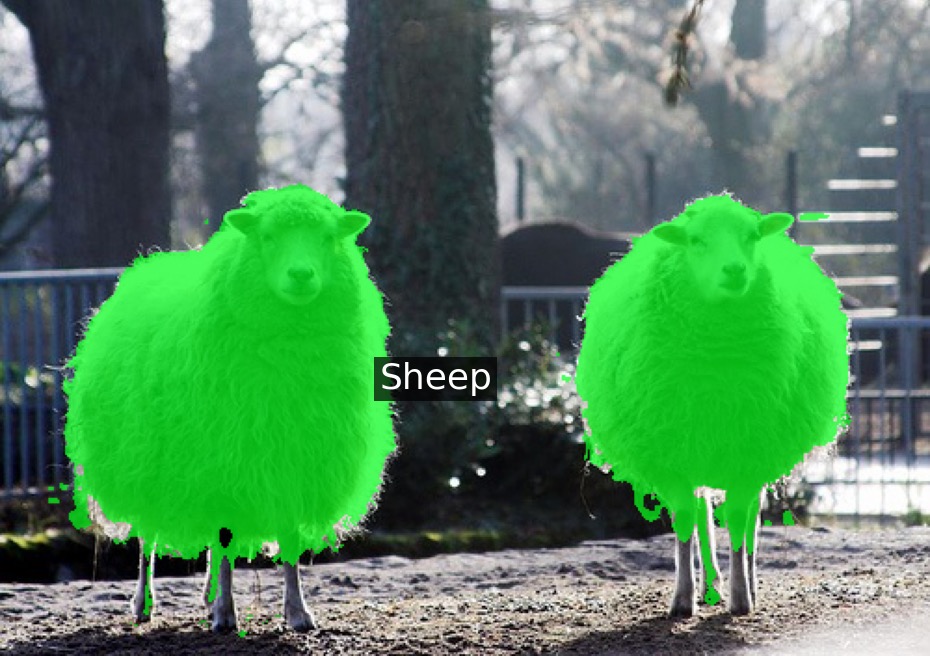} & \includegraphics[width=0.25\linewidth]{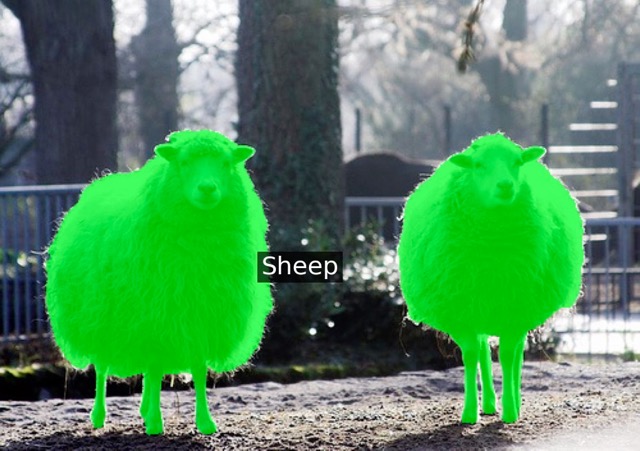}& \includegraphics[width=0.25\linewidth]{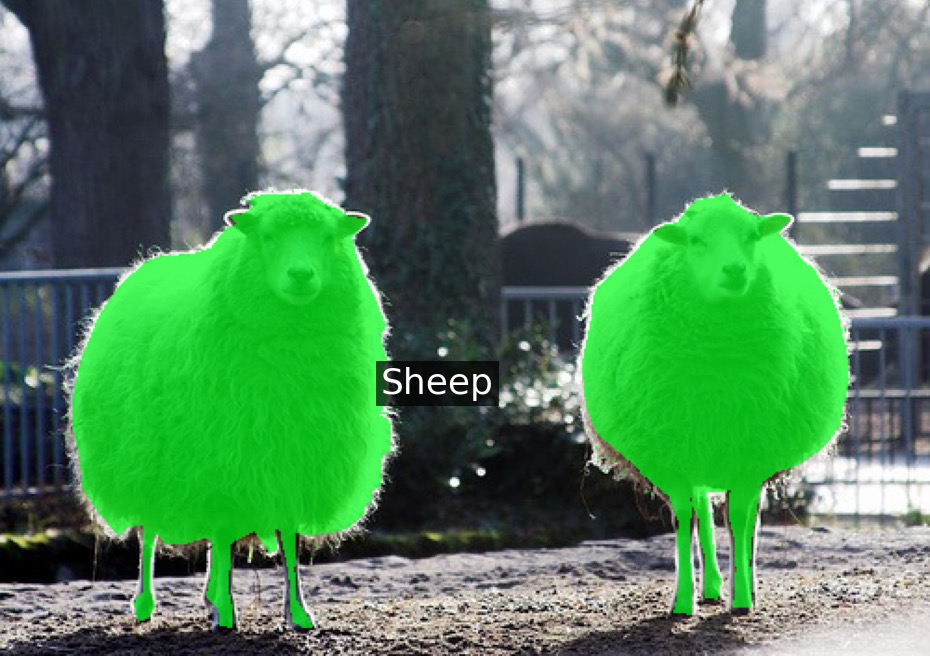} \\
      \includegraphics[width=0.25\linewidth]{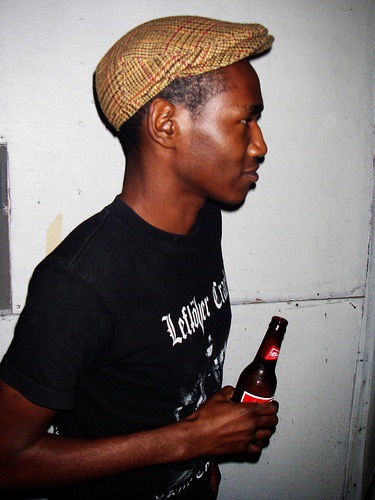} & \includegraphics[width=0.25\linewidth]{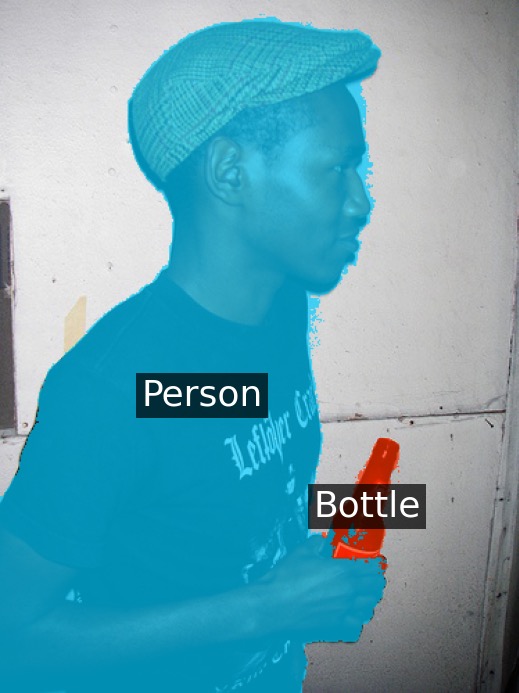} & \includegraphics[width=0.25\linewidth]{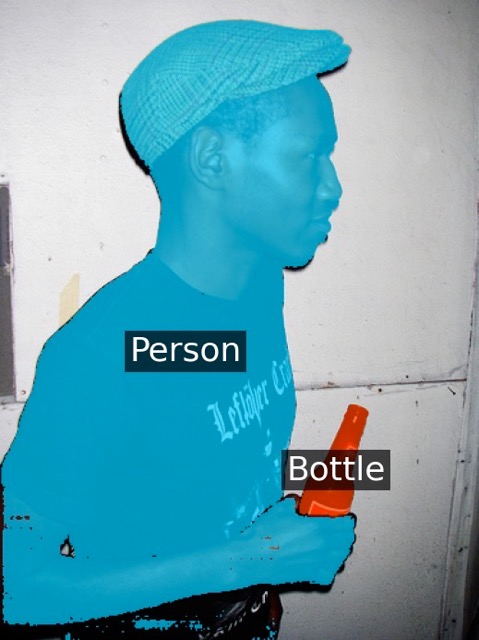}& \includegraphics[width=0.25\linewidth]{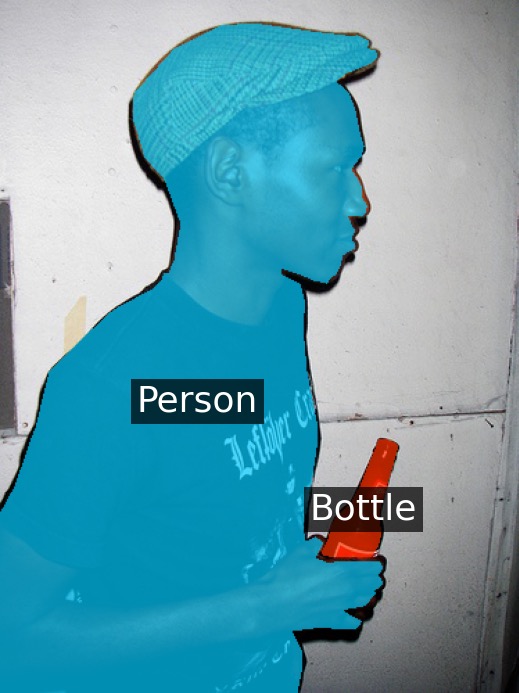} \\
      \includegraphics[width=0.25\linewidth]{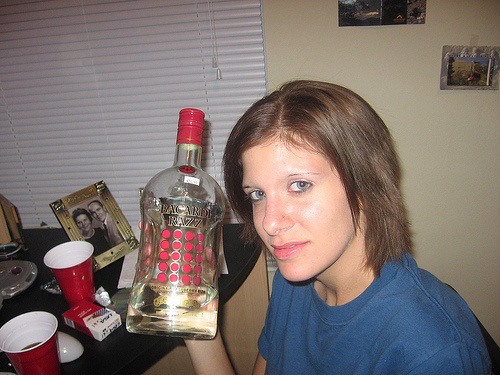} & \includegraphics[width=0.25\linewidth]{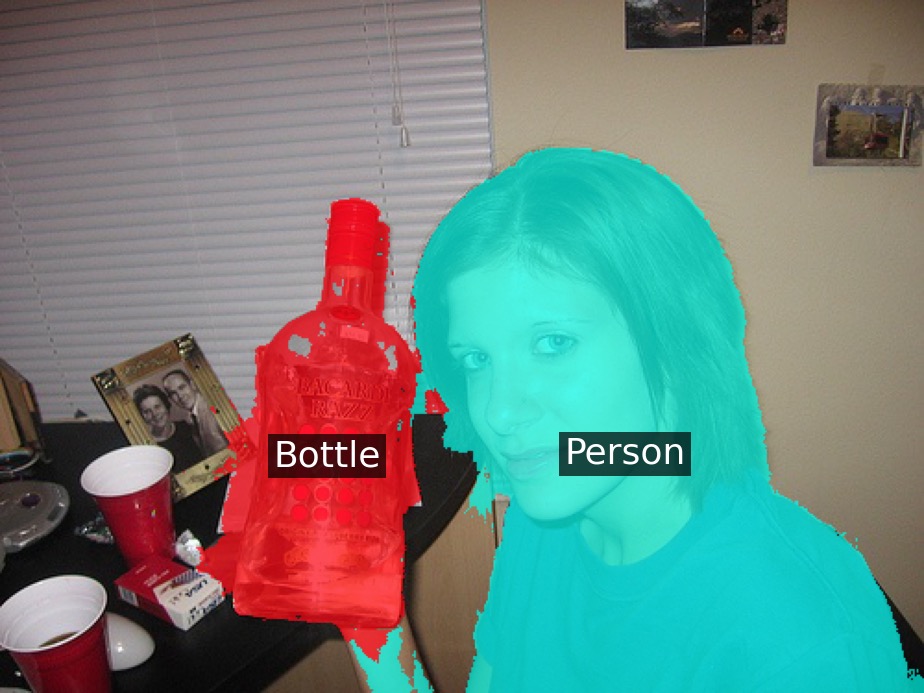} & \includegraphics[width=0.25\linewidth]{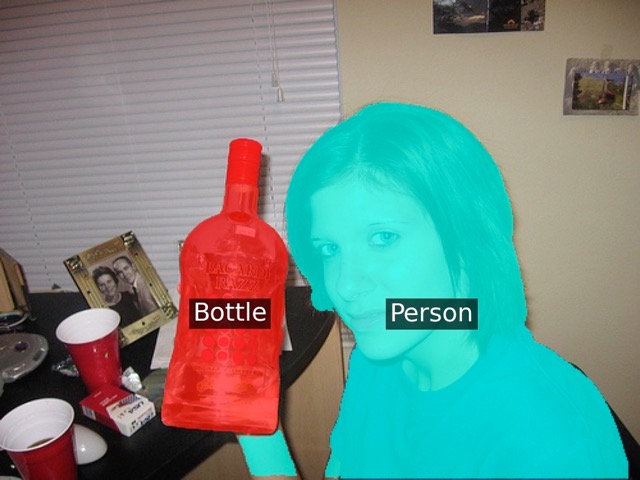} & \includegraphics[width=0.25\linewidth]{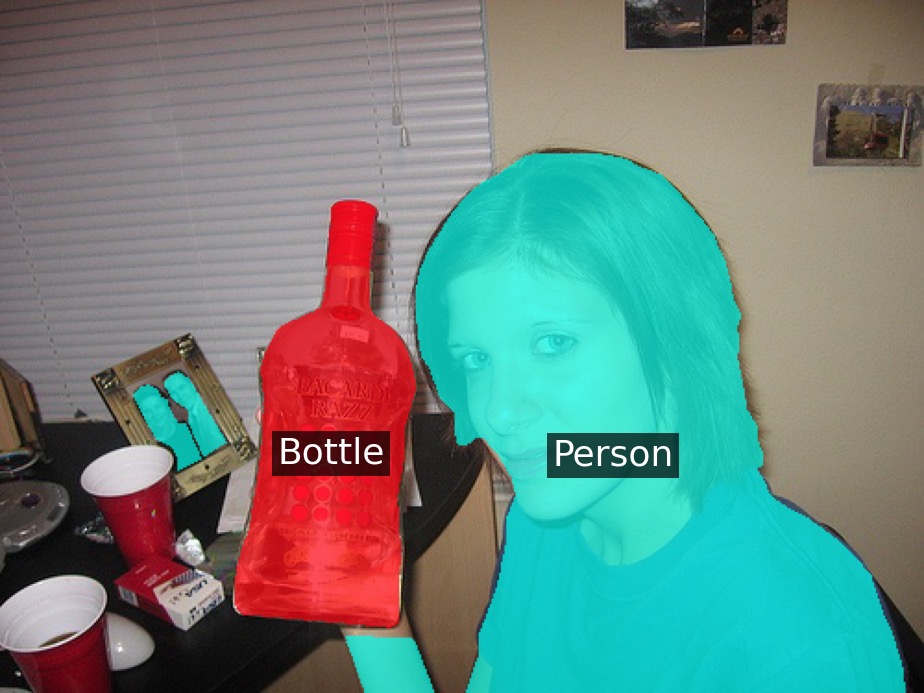}\\
    \end{tabular}
    \caption{Comparison of different post-processors on randomly selected images from PASCAL VOC.}
    \label{fig:post_compare_voc}
\end{figure*}

\begin{figure*}
    \centering
    \tablestyle{0pt}{0.1}
    \begin{tabular}{cccc}
      Image & \mname & \mname+SAM & GT \\
      \includegraphics[width=0.25\linewidth]{images/post_comparison/coco/original_images/0.jpg} & \includegraphics[width=0.25\linewidth]{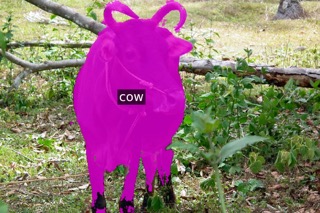} & \includegraphics[width=0.25\linewidth]{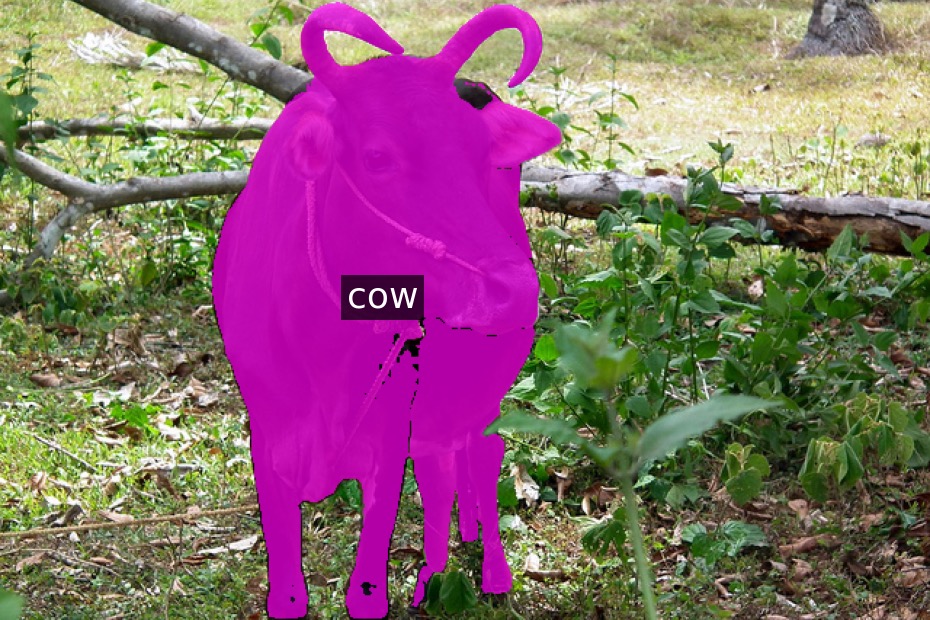} & \includegraphics[width=0.25\linewidth]{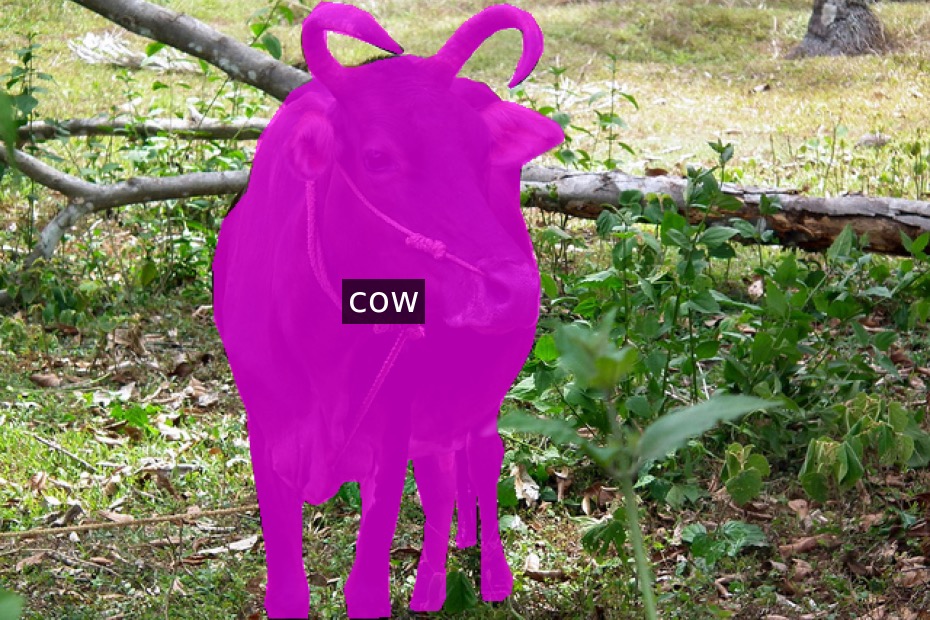}  \\
        \includegraphics[width=0.25\linewidth]{images/post_comparison/coco/original_images/1.jpg} & \includegraphics[width=0.25\linewidth]{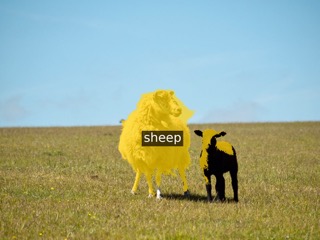} & \includegraphics[width=0.25\linewidth]{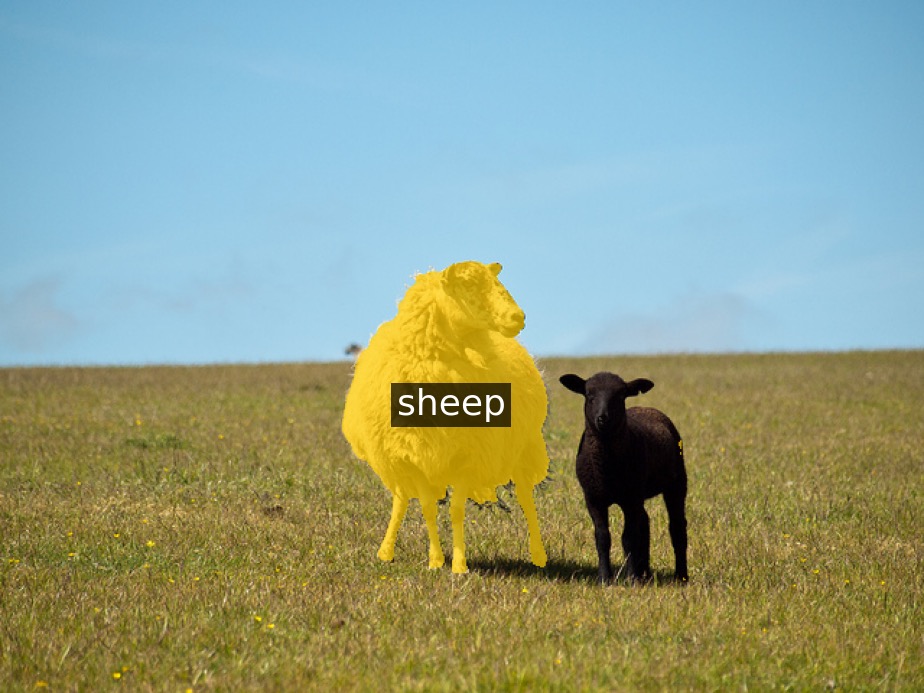}& \includegraphics[width=0.25\linewidth]{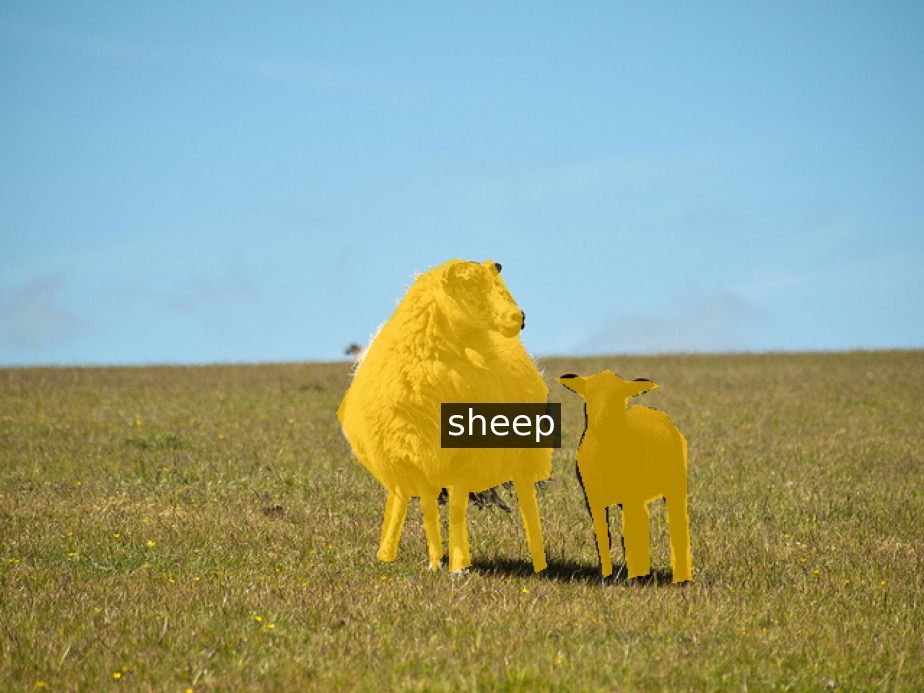} \\
      \includegraphics[width=0.25\linewidth]{images/post_comparison/coco/original_images/3.jpg} & \includegraphics[width=0.25\linewidth]{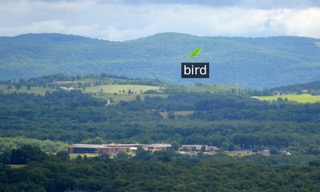} & \includegraphics[width=0.25\linewidth]{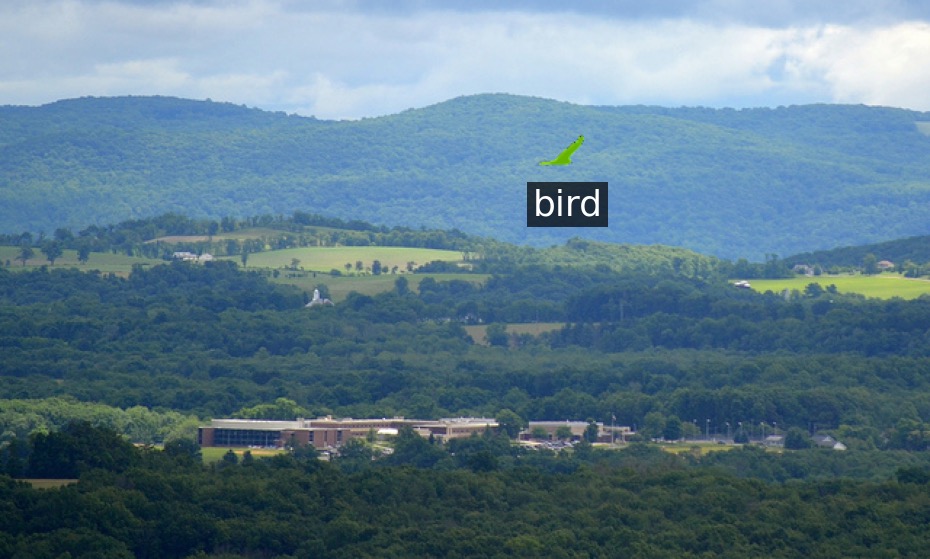}& \includegraphics[width=0.25\linewidth]{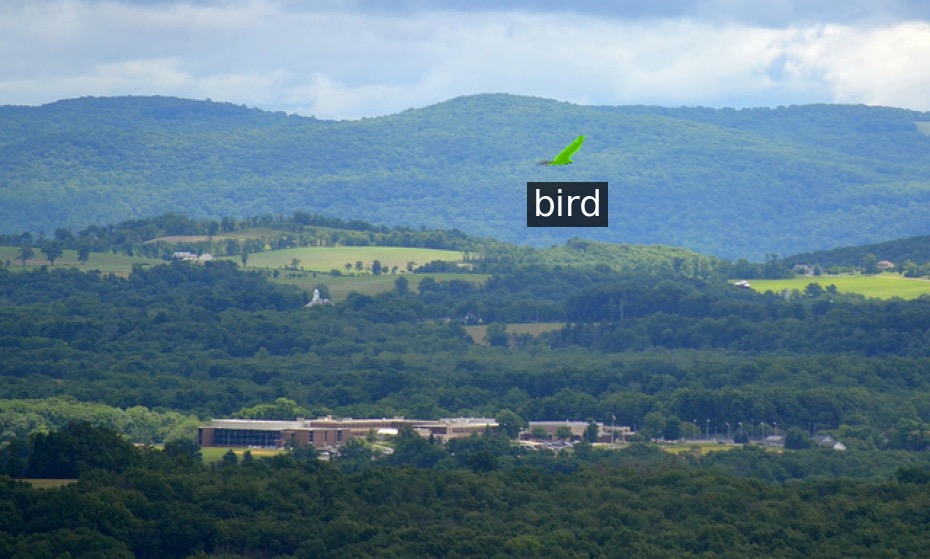} \\
      \includegraphics[width=0.25\linewidth]{images/post_comparison/coco/original_images/4.jpg} & \includegraphics[width=0.25\linewidth]{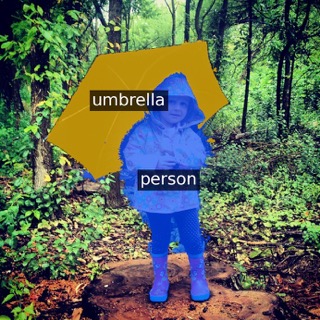} & \includegraphics[width=0.25\linewidth]{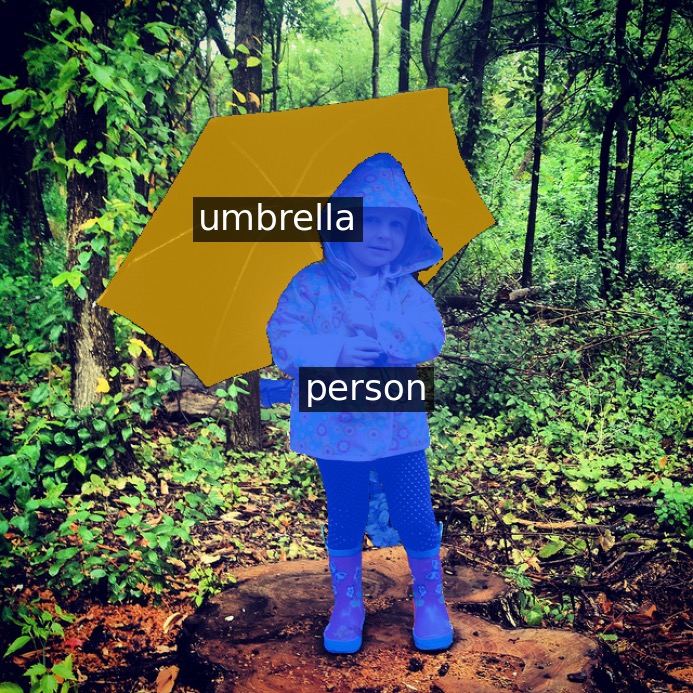}& \includegraphics[width=0.25\linewidth]{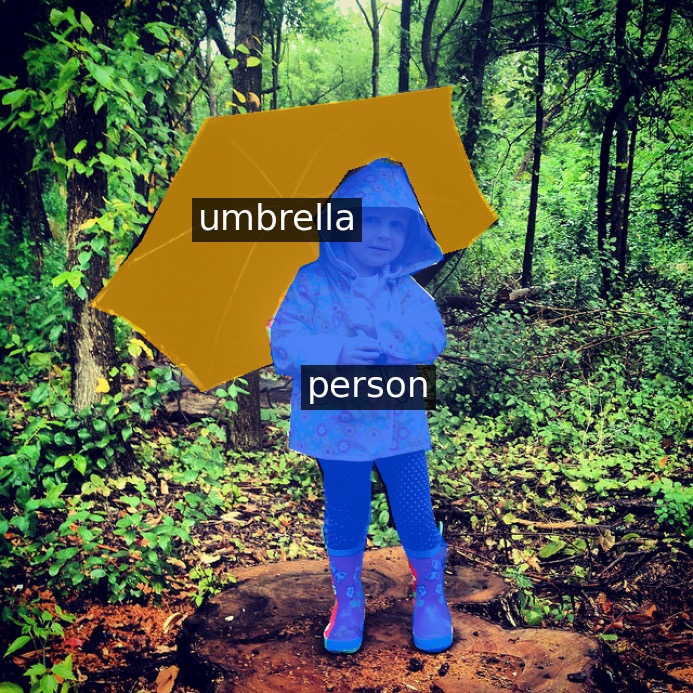} \\
      \includegraphics[width=0.25\linewidth]{images/post_comparison/coco/original_images/5.jpg} & \includegraphics[width=0.25\linewidth]{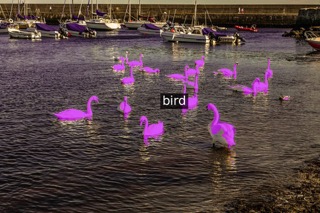} & \includegraphics[width=0.25\linewidth]{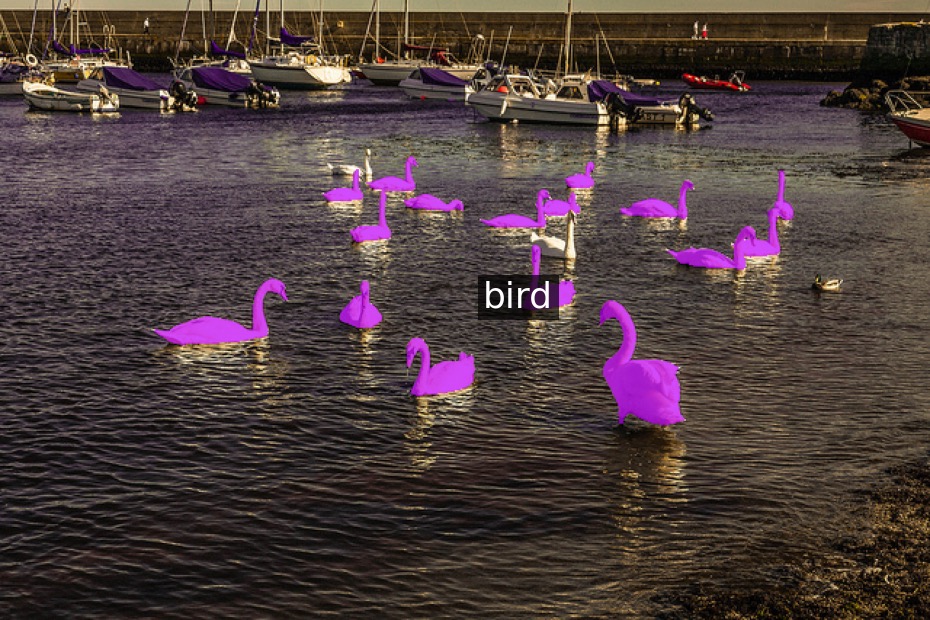} & \includegraphics[width=0.25\linewidth]{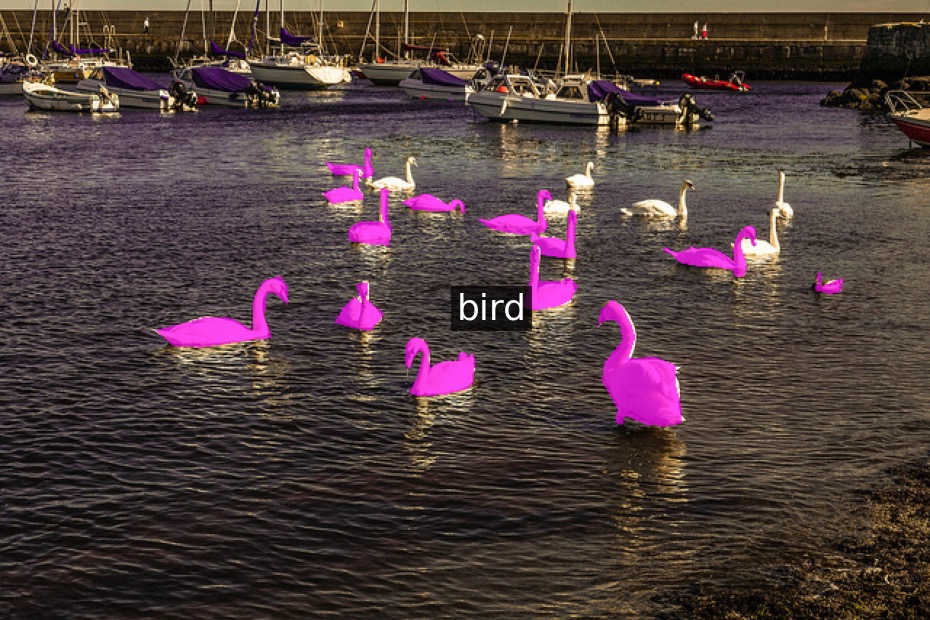}\\
    \end{tabular}
    \caption{Comparison of different post-processors on randomly selected images from COCO Object.}
    \label{fig:post_compare_coco}
\end{figure*}

Figures \ref{fig:post_compare_voc} and \ref{fig:post_compare_coco} present a comparative visualization of the post-processing techniques Conditional Random Field (CRF) and Segment Anything Model (SAM)~\cite{sam_hq}, applied to randomly chosen samples from the VOC~\cite{pascal_voc} and COCO Object datasets~\cite{coco_stuff}. Initial observations reveal that the application of CRF in \mname~facilitates the generation of high-quality masks, albeit with notable limitations in delineating boundaries between distinct semantic masks. The integration of SAM enhances the precision of these masks, yielding clearer and more well-defined boundaries. Nevertheless, the implementation of SAM is not without drawbacks; it occasionally leads to false negative predictions, stemming from mismatches between \mname~raw masks and SAM candidate masks (the matching algorithm is introduced in the main manuscript), or false positive predictions due to the overly coarse nature of SAM masks. Meanwhile, we find SAM is not very sensitive to stuff classes, so combining SAM on Pascal Context will not lead to much increase in mIoU.

\subsection{Visualization comparison for different open-vocabulary segmentation methods. }
\begin{figure*}
    \centering
    \tablestyle{0pt}{0}
    \begin{tabular}{p{0.25\linewidth}p{0.25\linewidth}p{0.25\linewidth}p{0.25\linewidth}}
     \makecell[c]{Image} & \makecell[c]{OVSeg~\cite{ovseg}} & \makecell[c]{Grounded SAM~\cite{liu2023grounding}} &  \makecell[c]{Ours}  \\  %
    \multicolumn{4}{c}{\includegraphics[width=\linewidth]{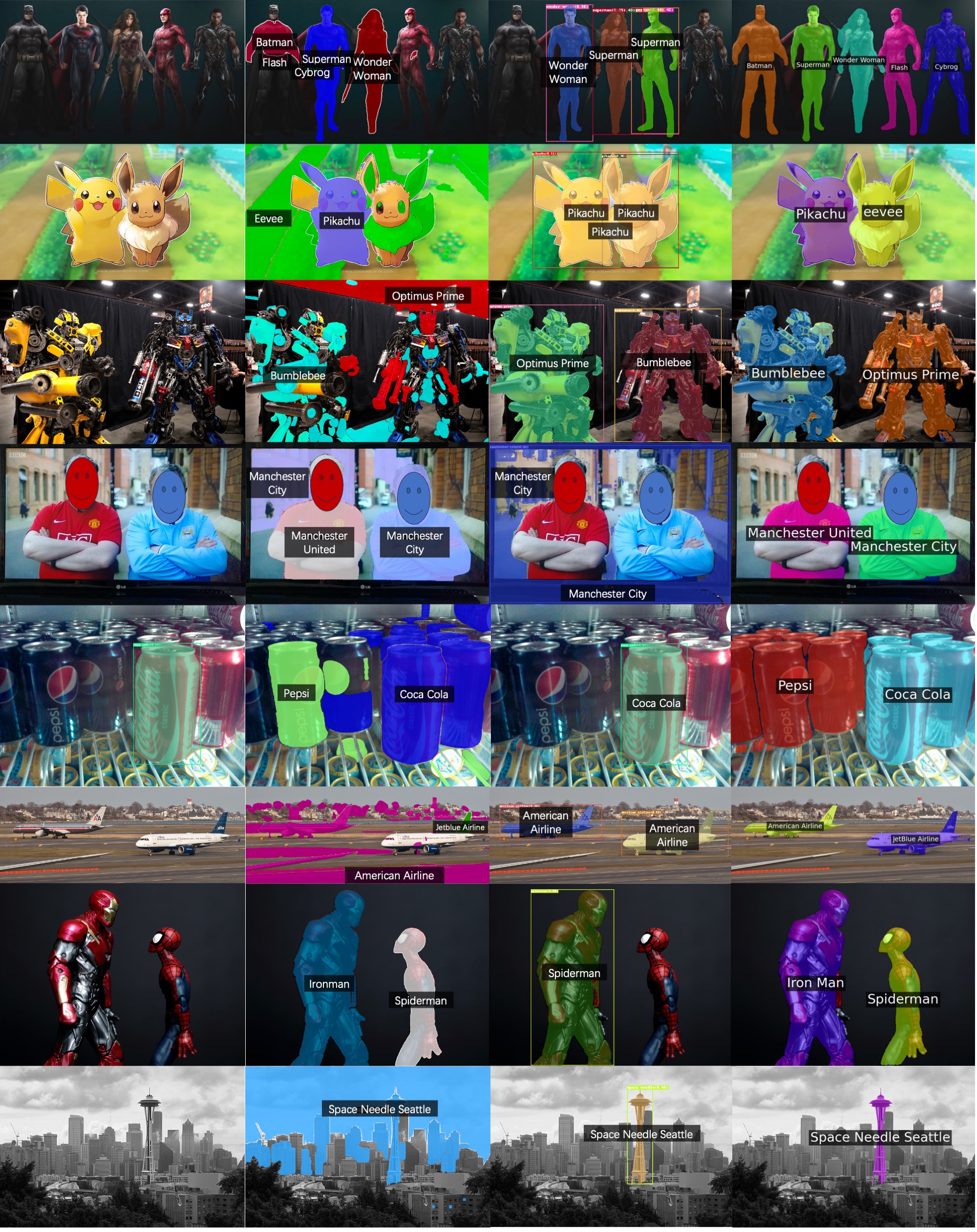}} \\
    \end{tabular}
    \caption{Visualization comparison of different open-vocabulary segmentation methods.}
    \label{fig:ov-compare}
\end{figure*}

Figure \ref{fig:ov-compare} presents a qualitative comparison of open-vocabulary segmentation results for a variety of non-standard subjects, including unique characters, brands, and landmarks. These subjects are notably distinct from common objects. The Grounded SAM~\cite{liu2023grounding} method demonstrates proficiency in segmenting prominent objects with precision, yet it often misclassify these segments. The OVSeg~\cite{ovseg} approach also generates low-quality segmentation masks and inaccurate class predictions. In contrast, our methodology \mname{} excels by creating high-quality masks with accurate semantic class predictions, showcasing its superior capability in the realm of open-vocabulary segmentation.

\section{Limitation}
The primary limitation of our method is that its performance is bounded by the pre-trained VLM. For example, since the CLIP model utilizes horizontal flipping augmentation during training, it becomes challenging for our model to successfully distinguish between the concepts ``left" and ``right". However, we believe that this issue can be easily resolved through adjustments, such as incorporating better data augmentation techniques during the pre-training phase.

\section{Future Potentials and Broader Impact}
\mname{} is simple, straightforward yet highly efficient. To enhance its performance further, we provide two ways to explore. First, incorporating additional trainable modules such as Feature-Pyramid Networks can significantly improve its capability in handling small objects. Second, since our method is fundamentally compatible with various Vision-Language Models (VLMs), it presents an intriguing opportunity to investigate integration with other VLMs. Moreover, \mname{} can serve the purpose of generating pseudo-labels for other open-vocabulary segmenters.

\end{document}